\documentclass[twocolumn]{autart}    

\usepackage{graphicx}          
\usepackage[dvips]{epsfig}    
\usepackage{booktabs}
\usepackage{multirow}                           
\usepackage{epstopdf}
\usepackage{subcaption}
\usepackage{bm}
\usepackage{fontspec}
\usepackage{color}
\usepackage{stfloats}
\usepackage{amsmath,amsfonts} 
\usepackage{mathrsfs}
\newtheorem{assumption}{Assumption}
\newtheorem{remark}{Remark}
\newtheorem{definition}{Definition}
\newtheorem{corollary}{Corollary}
\newtheorem{theorem}{Theorem}

\usepackage{algorithmic}
\usepackage{algorithm}
\usepackage{hyperref}

\newcommand{\diam}[1]{\mathrm{diam}{#1}}

\allowdisplaybreaks[4]
\begin{document}

\begin{frontmatter}

\title{Receding Fixed-Horizon Optimization for Near-Time-Optimal Trajectory Planning and Control\thanksref{footnoteinfo}} 

\thanks[footnoteinfo]{This work is supported in part by the National Natural Science Foundation of China (62173191)}

\author{Haotian Tan}\ead{ElmundoTan@outlook.com},    
\author{Yuan-Hua Ni}\ead{yhni@nankai.edu.cn}              

\address{College of Artificial Intelligence, Nankai University, Tianjin 300350, PR China}                                     


\begin{abstract}                          
Time-optimal trajectory planning and control is central for autonomous vehicles, yet its application and real-time deployment confronts two fundamental challenges: the non-convexity of optimal control problems and the unpredictable computation time inherent to nonlinear programming. To address these challenges, we propose a hierarchical convex optimization framework that addresses both issues by decomposing the original problem into short, fixed-horizon planning cycles. Each cycle solves a convex subproblem within a collision-free region identified by a customized search algorithm; the complete trajectory and control is assembled by concatenating state-input sequences across cycles. Under mild assumptions, we establish finite-time convergence of the decomposition procedure and show that the concatenated solution satisfies the necessary conditions for local optimality. Numerical experiments on randomly generated maps with static and dynamic obstacles demonstrate that the proposed algorithm achieves a higher success rate and substantially lower computation time than sequential convex programming, while maintaining comparable control time. These results show that decomposition-based convex optimization provides a practical pathway to reliable, real-time near-time-optimal trajectory planning.
\end{abstract}

\end{frontmatter}

\section{Introduction}
Trajectory planning for autonomous vehicles seeks trajectories or control inputs that steer the vehicle from its initial state to a target region while avoiding obstacles. Algorithms in this area are commonly classified into search-based, sampling-based, and optimization-based methods \cite{three_methods}. Search-based planners such as A* \cite{A*} use heuristics to guide a search over a discretized environment, but scale poorly with environment size. Sampling-based planners such as RRT \cite{RRT} and its asymptotically optimal variant RRT* \cite{RRT*} are simple and flexible, yet the paths they produce are often not dynamically feasible. For both classes, the computation time is difficult to bound, which complicates real-time operation.

Motion planning can be posed more faithfully as an optimal control problem, in which a cost functional, such as travel time or energy consumption, is minimized subject to the vehicle dynamics and to state and input constraints. Numerical methods for such problems follow two philosophies. Indirect methods invoke Pontryagin's maximum principle (PMP) or Lagrange multiplier theory to characterize the optimal solution through necessary conditions, which are then solved numerically as two-point boundary-value problems \cite{Pon,Lag}. These methods offer structural insight but are sensitive to initialization and become difficult to apply when state constraints or active inequality constraints are numerous. Direct methods transcribe the continuous-time problem into a finite-dimensional nonlinear program (NLP) over the discretized states and controls \cite{rao2009,betts2010}, which can then be solved by Newton-type iterations or by operator-splitting schemes such as the alternating direction method of multipliers \cite{Newton,ILQR,ADMM0}. Direct transcription is robust and widely used, but the resulting NLP is non-convex in general \cite{nocedal2006}, so the solution time of a generic NLP solver is unpredictable; for onboard real-time applications this unpredictability is a serious drawback \cite{tutorial}.

Convex optimization removes this unpredictability for structured classes of optimal control problems \cite{boyd2004}. When the non-convexity lies in the control constraints, lossless convexification produces a convex problem equivalent to the original one; this has been established for powered-descent guidance under thrust and pointing constraints \cite{LCVX1,LCVX2,LCVX3} and generalized to a class of nonlinear optimal control problems \cite{lsch}. When the non-convexity lies in the state constraints, as in obstacle avoidance, sequential convex programming (SCP) iterates between linearizing the dynamics and constraints about the current trajectory and solving the resulting convex subproblem within a trust region \cite{SCP,GuSTO,TRSCVX}, and has been demonstrated for quadrotor fleets, spacecraft, and autonomous racing \cite{SCP,TRSCVX,Racing}. Nevertheless, SCP typically converges to a locally sub-optimal solution, and the number of iterations, hence the computation time, is not deterministically bounded.

Time-optimal control, the subject of this paper, adds a distinct difficulty: the final time is itself a decision variable, so the problem cannot be posed on a fixed time grid. The maximum principle provides necessary conditions for minimum-time solutions, whose structure is bang-bang for broad classes of linear systems \cite{pontryagin1962}; numerically, however, the problem is non-convex in both the trajectory and the horizon. In discrete time the horizon length $N$ is an additional integer decision, and the optimal final time is an integer multiple of the discretization step. One approach is to search over $N$, solving a fixed-horizon problem for each candidate; alternatively, time-optimal trajectories can be computed from a sequence of fixed-time subproblems with progressively adjusted horizon, as in autonomous racing \cite{Racing}. Such schemes, however, typically require a complete map, which may limit their applicability in unknown and dynamic environments. Receding-horizon model predictive control (MPC) addresses the limitation: a short-horizon problem is re-solved at each step and only the first control is applied, so the per-step computation is bounded and the plan adapts to newly observed obstacles \cite{mayne2000}. Recent work has exploited receding-horizon planning for near-time-optimal flight, e.g., aggressive quadrotor maneuvers \cite{foehn2021,romero2022}. Attaining time-optimality within a receding-horizon framework nonetheless remains difficult: the horizon must be kept short enough to preserve predictable per-step computation, yet long enough to recover the time-optimal horizon, and terminal conditions that certify optimality are generally absent in unknown, dynamic environments.

This paper proposes a receding fixed-horizon convex optimization framework for collision-free, time-optimal trajectory planning of autonomous vehicles in environments with both static and dynamic obstacles. Instead of solving a single large non-convex NLP, the framework decomposes the problem into a sequence of short, fixed-horizon planning cycles. Each cycle proceeds in three steps. First, a customized trajectory search algorithm generates a dynamically feasible reference trajectory and extracts from the map a collision-free convex region around it. Second, a relaxed convex subproblem, obtained by omitting the obstacle constraints, yields a nominal trajectory. Third, a strictly convex subproblem, augmented with a safety term, is solved within this region to produce the trajectory segment of the cycle. The segments are concatenated across cycles into a globally admissible trajectory. Because every subproblem is convex and of short, fixed horizon, each cycle is solved by a standard interior-point method in a predictable, essentially constant amount of time, and re-planning relies only on local map information, which enables operation in dynamic environments.

The trajectory produced by the framework is not exactly time-optimal in general. It is certified to satisfy the Karush-Kuhn-Tucker (KKT) necessary conditions of the discretized linearized subproblem, provided that a constraint qualification such as the Mangasarian-Fromovitz condition holds \cite{MFCQ}. The deviation from local optimality is small and stems from three factors inherent to discrete-time, model-based planning: the final time is quantized to an integer multiple of the discretization step; the dynamics are approximated by an affine model within each cycle; and the preconditions of the local-optimality guarantee occasionally fail, with limited impact on the final time in our numerical experiments.

The main works and contributions of this paper are summarized as follows.
\begin{itemize}
	\item We propose a planning‑cycle framework that reformulates the time‑optimal NLP as a sequence of convex subproblems with short, fixed horizons, each solved over a collision‑free convex region extracted by trajectory search. Of the two trajectory‑search strategies presented, our improved vortex artificial‑potential‑field method constitutes an original contribution: it generates dynamically feasible reference trajectories for the proposed framework and can function as a standalone planner in its own right. By contrast, a customized adaptation of the dynamic‑window approach \cite{DWA} is also included as an alternative option for general nonlinear vehicle models.
	\item Under mild assumptions, we prove that the planning-cycle procedure terminates after finitely many cycles and that the concatenated trajectory satisfies the KKT necessary conditions of the linearized subproblem; candidate local optimality is then established.
	\item Numerical experiments on randomly generated maps with static and dynamic obstacles show that the proposed method achieves a higher success rate and substantially lower computation time than sequential convex programming, at a comparable final time.
\end{itemize}

In the remainder of this paper, Section 2 formulates the time-optimal control problem, introduces the fixed-horizon subproblem and its linearization, and describes the planning-cycle structure. Section 3 presents the proposed framework, including the convex reformulation, the complete algorithm, the two trajectory search algorithms, and the theoretical analysis. Section 4 reports numerical experiments in static and dynamic environments. Section 5 concludes the paper.

\textit{Notations:} throughout the subsequent sections, all sets are defined in $\mathbb{R}^n$ with context-specific $n$, endowed with the Euclidean norm $\left\| {\mathbf{a}} \right\| = \sqrt {\left\langle {\left. {{\mathbf{a,a}}} \right\rangle } \right.}  = \sqrt {{{\mathbf{a}}^ \top }{\mathbf{a}}} $, and we further let $\|\mathbf a \|_Q = \sqrt{\mathbf a^\top Q \mathbf a}$ with $Q$ being a positive definite weighting matrix. 
The diameter of a set $\mathcal{S} \subset \mathbb{R}^n$ is defined as
\begin{equation*}
	{\diam{(\mathcal S)}} := \left\{ \begin{array}{l}
		\quad	0, \hfill \mathcal S = \emptyset ,\\
		\sup \{ \|\mathbf z - \mathbf x\| :\mathbf x,\mathbf z \in \mathcal S\}, \hfill \quad else,
	\end{array} \right.
\end{equation*}
and for some $\mathbf x_0 \in \mathbb R^n$ with non-negative real number $d$, we denote by $\mathcal B(\mathbf x_0, d)$ the closed ball centered at $\mathbf x_0$ with radius $d$:
\[\mathcal B(\mathbf x_0, d) := \{\mathbf x \in \mathbb R^n : \|\mathbf x - \mathbf x_0\| \leq d\}.\]
Sliced vectors are denoted $\mathbf{a}[i:j]$, representing rows $i$ to $j$ of column vector $\mathbf{a}$ (e.g., $\mathbf{a} = [a_1,a_2,a_3,a_4]^\top \Rightarrow \mathbf{a}[2:3] = [a_2,a_3]^\top$). For a set $\mathcal{S} \subseteq \mathbb{R}^{n_s}$, the sliced set is defined as
\[\mathcal S[a:b] := \big\{ \mathbf x[a:b] \big| \mathbf x \in \mathcal S \big\}.\]
The gradient of a differentiable function $f : \mathbb R^n \to \mathbb R$ is denoted $\nabla f(\mathbf{x}) := \left[ \frac{\partial f}{\partial x_1}, \cdots, \frac{\partial f}{\partial x_n} \right]^\top$. We use $\mathrm{Null}(\cdot)$ for the null space, and $\mathrm{cl}(\cdot)$ for the closure. 

\section{Preliminaries}
\subsection{Problem Formulation}
\begin{figure}[h]
	\centering
	\includegraphics[width=0.85\linewidth]{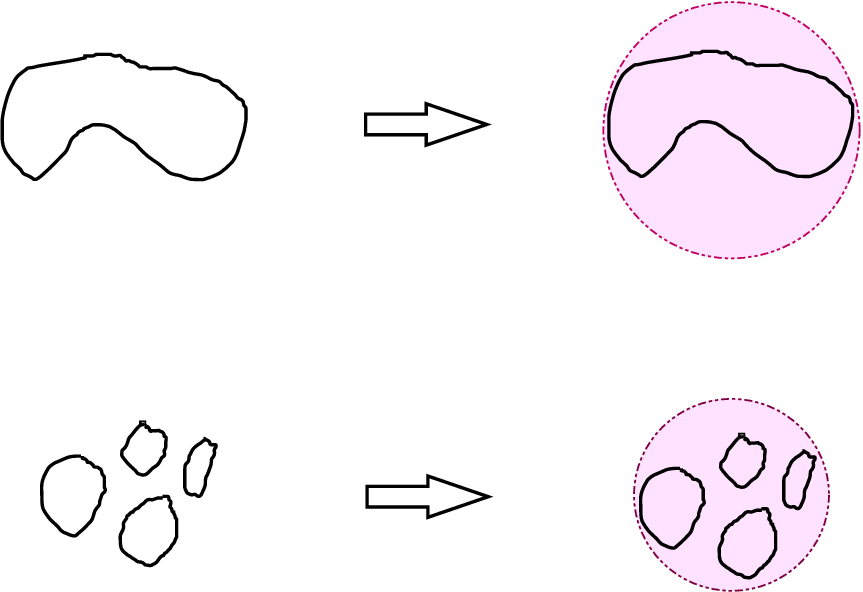}
	\caption{Spherical approximation of obstacle shapes: an irregular obstacle enclosed by a single ball (top) and several clustered small obstacles combined into a larger enclosing ball (pink, bottom right).}
	\label{fig:fig1}
\end{figure}

We consider the problem of generating a collision-free trajectory for a 2-D vehicle from an initial state to a target region within a structured environment $\mathcal{M}$ (the map), defined in unified Euclidean space. Let $\mathcal{O}_i$, $i \in \{1,\cdots,n_{obs}\}$, denote the obstacles, each bounded since $\mathcal{M}$ is bounded. For irregularly shaped obstacles, we adopt a conservative spherical approximation: each $\mathcal{O}_i$ is enclosed by a closed ball $\mathcal{B}(p_i,\diam{(\mathcal{O}_i)}/\sqrt{3})$, where $x$ and $p_i$ denote positions of the vehicle and obstacles in $\mathcal{M}$, and the vehicle is treated as a point with its clearance from obstacle $i$ enlarged by a constant margin. Denoting $\mathcal{B}_i:=\mathcal{B}(p_i,\diam{(\mathcal{O}_i)}/\sqrt{3})$ and taking $r_i>0$ as the safe separation radius around $p_i$, the obstacle avoidance constraint becomes
\begin{equation}
	\|x - p_i\| \ge r_i, \label{nonconvex_position_con}
\end{equation}
eliminating the need for explicit shape knowledge. This approximation relies on two assumptions.
\begin{assumption}
	The initial and target positions lie outside all $\mathcal B_i$, $i \in \{1,\cdots,n_{obs}\}$. \label{original_A1}
\end{assumption}
\begin{assumption}
	The obstacle regions are mutually disjoint. Moreover, there exists $\ell > 0$ such that $\mathcal B(p_i,r_i + 0.5\ell) \cap \mathcal B(p_j,r_j + 0.5\ell) = \emptyset$ for all $i \neq j$. \label{original_A2}
\end{assumption}
Assumption~1 underlies all subsequent analysis. Assumption~2 allows multiple nearby small obstacles to be treated as a single larger obstacle (Fig.~\ref{fig:fig1}) and is therefore mild.

While $x \in \mathbb R^2$ denotes the 2-D position of the vehicle, further pose coordinates, such as heading angle, can be incorporated as states. We collect these coordinates into the pose vector $x_+ \in \mathbb R^{n_+}$, whose first two entries are the 2-D position, $x = x_+[1:2]$. Assuming the pose parameters to be twice differentiable with respective to $t$, we take the state of the vehicle as $z(t) := [x_+^\top,\dot x_+^\top]^\top$ and describe its dynamics by
\begin{equation}
	\dot{z}(t) = f_c(z,u), \label{non_linear_original}
\end{equation}
with differentiable $f_c : \mathbb R^{n_z} \times \mathbb R^{n_u} \to \mathbb R^{n_z}$, where $u$ is the control input.

The double-integrator model, for which $f_c$ yields $x_+ = x$ with physical acceleration as control input, is a representative member of the class of vehicle dynamics considered in this paper. Crucially, both the formulation and our proposed algorithm apply to a broader range: for the unicycle or Ackermann model, we just need to substitute $x$ by $[x_x,x_y,\theta]^\top$ with $u=[v,\omega]^\top$; while for the planar quadrotor dynamics, take $x=[x_x,x_z,\phi]^\top$ with thrust and torque as inputs. The same planning-cycle framework generalizes directly via the unified dynamics $z_{i+1}=f(z_i,u_i)$.

Physical constraints are encoded as $z^\top Q_1 z \le 1$ and $u^\top Q_2 u \le 1$, where $Q_1,Q_2$ are strictly positive definite. These subsume generalized speed limits ($\|\dot x\| \leq v_{\max}$), input bounds ($\|u\|\leq u_{\max}$), etc. Let $z_0$ and $z_f$ denote the initial and a target state respectively. Boundary constraints are
\begin{align}
	& z(0) = z_0, \label{initial_condition} \\
	& \|z(t_f) - z_f\|_{Q_0} \leq \gamma, \label{boundary_condition}
\end{align}
where $Q_0$ is a positive definite weighting matrix and $t_f$ the final time. With aforementioned notations, a time-optimal trajectory optimization problem can be constructed as
\begin{alignat}{2}
	& \min_{z(t),u(t)} && \; t_f \\
	& \quad \ \ \mathrm{s.t.} && \; \eqref{non_linear_original} - \eqref{boundary_condition}, \\
	& && \; \|z(t)\|_{Q_1}^2 \leq 1,\quad \|u(t)\|_{Q_2}^2 \leq 1.
\end{alignat}
For numerical solution, discretize with step $h$, substituting $z(t)$ by $\{z_1,\dots,z_{N+1}\}$ and $u(t)$ by $\{u_1,\dots,u_N\}$. The dynamics become
\begin{align}
	&z_{i+1} - z_i = h f_c(z_i,u_i) \Rightarrow\; z_{i+1} := f(z_i,u_i). \label{c2d_original}
\end{align}
Denote $\mathbf y_N := \{z_1,\dots,z_{N+1}; u_1,\dots,u_N\}$. The discretized problem $\mathcal P_0$ reads
\begin{align}
	& \min_{\mathbf y_N} &&Nh \nonumber \\
	& \quad \mathrm{s.t.} &&\eqref{c2d_original}, \nonumber\\
	& &&G_1(z_i) := z_i^\top Q_1 z_i - 1 \leq 0, \label{DQ_1} \\
	& &&G_2(u_i) := u_i^\top Q_2 u_i - 1 \leq 0, \label{DQ_2} \\
	& &&z_1 = z_0, \label{initial_condition_discrete} \\
	& &&\|z_{N+1} - z_f\|_{Q_0} \leq \gamma, \label{boundary_condition_d}\\
	& &&\|z_i[1:2] - p_j\| \ge r_j,\; 1 \leq j \leq n_{obs}. \label{nonconvex_position_con_discrete}
\end{align}

Problem $\mathcal P_0$ appears widely in a variety of trajectory or motion planning tasks. However, it couples non-convex obstacle avoidance constraints \eqref{nonconvex_position_con_discrete} with a time-minimizing objective over a variable horizon $N$, forming a challenging non-convex NLP. Standard nonlinear solvers applied to such problems offer no guarantee of solution quality and exhibit unpredictable computation times, which may render them unsuitable for real-time onboard deployment. These limitations motivate a decomposition framework that produces near-optimal trajectories with reliable, deterministic computation --- the central challenge addressed in this work.

\subsection{Linearized Subproblems and Planning Cycles}
We assume $\mathcal P_0$ is solvable, a mild condition typical in related literature. Rather than solving $\mathcal P_0$ directly, our framework iteratively constructs and solves discretized subproblems with a fixed time step $h$ and a preselected horizon $N$, then concatenates the resulting trajectory segments.

To construct subproblems, a reference state, which is the initial state of the corresponding subproblem, is required. For a reference state $z_0'$ and reference control $u'_0$, we linearize \eqref{c2d_original} at the single state-input pair $y_0 := (z'_0,u'_0)$, which gives
\begin{equation}
	z_{i+1} := A(y_0) z_i + B(y_0) u_i + w(y_0), \label{c2d_var}
\end{equation}
where $A(y_0), B(y_0)$ are the Jacobians of $f$ at $y_0$ and $w(y_0)$ the offset; for simplicity, we denote $(z'_0, u'_0)$ by $y_0$ throughout the rest of the paper. Especially, for the double-integrator case ($u$ = acceleration),
\[
A(y_0) \equiv \begin{pmatrix} \mathbf I_{2} & h\mathbf I_{2}\\ \mathbf O_{2} & \mathbf I_{2} \end{pmatrix},\quad
B(y_0) \equiv \begin{pmatrix} \mathbf O_{2} \\ h\mathbf I_{2} \end{pmatrix},\quad
w(y_0) \equiv \mathbf 0.
\]
Define the cost of a state-input sequence $\mathbf y_N$ by
\begin{equation}
	J(\mathbf y_N) = [z_f - z_{N+1}]^\top Q_0 [z_f - z_{N+1}], \label{J_P_1}
\end{equation}
then, for a given reference pair $y_0 = (z'_0,u'_0)$, two problem classes can be further defined. The first, $\mathcal P_1(z'_0; N)$:
\begin{align}
	&\min_{\mathbf y_N} \quad \eqref{J_P_1} \nonumber \\
	&  \ \ \ \mathrm{s.t.}     \ \ \ \eqref{c2d_original} - \eqref{DQ_2},\; \eqref{nonconvex_position_con_discrete}, \nonumber \\
	&           \quad \quad \quad  z_1 = z'_0, \label{new_init}
\end{align}
and its linearized variant $\mathcal P_1'(y_0; N; A(y_0), B(y_0))$:
\begin{align}
	\min_{\mathbf y_N} \quad & \eqref{J_P_1} \nonumber \\
	 \mathrm{\; s.t.} \quad      & \eqref{DQ_1},\; \eqref{DQ_2},\; \eqref{nonconvex_position_con_discrete},\; \eqref{c2d_var},\; \eqref{new_init}. \nonumber
\end{align}
We use simplified notations $\mathcal P_1(\cdot), \mathcal P_1'(\cdot)$ when clear from context. Here $N$ is the prediction horizon and $Nh$ the control time. The original problem $\mathcal P_0$ is solved by iteratively solving $\mathcal P_1(\cdot)$ with small $Nh$. For weakly nonlinear systems with sufficiently short control horizons, the linearized dynamics with continuous updates to $A(y_0), B(y_0), w(y_0)$ preserve accuracy: the solution of $\mathcal P_1'(y_0; N; A(y_0), B(y_0))$ closely approximates that of $\mathcal P_1(z'_0; N)$. Hence, each short-horizon problem can be replaced by its linearized counterpart, with $y_0, N, A(y_0), B(y_0)$ updated in real time. The formulation and solution of each $\mathcal P_1(\cdot)$ (or its linearized form) constitutes one \emph{planning cycle}. 
\section{Proposed Solution Framework}
This section details the sequential subroutines that constitute each planning cycle: solving a relaxed problem to obtain a nominal solution, extracting a collision-free convex region, and finally solving a strict convex optimization problem within this region to obtain the trajectory segment of the cycle. Corresponding mathematical analyses are conducted to validate the efficacy of the framework.
\subsection{Reformulation of Planning Cycles}
With given $y_0, N, A(y_0), B(y_0)$, we denote the relaxed problem by $\mathcal P_r(y_0; N; A(y_0), B(y_0))$, which is derived from $\mathcal P_1'(y_0; N; A(y_0), B(y_0))$ while all the non-convex obstacle avoidance constraints \eqref{nonconvex_position_con_discrete} are removed, i.e. we have $\mathcal P_r(y_0; N; A(y_0), B(y_0))$:
\begin{align}
	&\min_{\mathbf y_N} && \eqref{J_P_1} \nonumber \\
	&\quad \mathrm{s.t.} && \eqref{DQ_1},\; \eqref{DQ_2},\; \eqref{c2d_var},\; \eqref{new_init}. \nonumber
\end{align}
The convexity of $\mathcal P_r(\cdot)$ is ensured since \eqref{nonconvex_position_con_discrete} is not considered in this problem. If $\mathcal P_r(y_0; N; A(y_0), B(y_0))$ is solvable, then its optimizer is referred to as the nominal solution of $\mathcal P_r(y_0; N; A(y_0), B(y_0))$.
\begin{remark}
	Note that we have $z_1 = z'_0$ according to \eqref{new_init}. Therefore, $\mathcal P_1(z'_0; N)$ and $\mathcal P_1'(y_0;N;\cdot)$ can also be viewed as optimization problems of $\{z_2, \cdots, z_{N + 1}, u_1, \cdots, u_N\}$.
\end{remark}

Suppose $\tilde{\mathbf y}_N$ is a feasible state-input sequence of $\mathcal P_1'(y_0; N; \allowbreak A(y_0), B(y_0))$ which is defined by
$$\tilde{\mathbf y}_N := {\{\tilde z_1, \cdots ,\tilde z_{N+1}; \tilde u_1, \cdots , \tilde u_N\}},$$ and $\bar{\mathbf y}_N := \{\bar z_1, \cdots ,\bar z_{N+1}; \bar u_1, \cdots , \bar u_N\}$ is one optimizer of the corresponding $\mathcal P_r(y_0; N; A(y_0), B(y_0))$, then a linear interpolation can be implemented as
\begin{align*}
	z_i[1:2] = (\lambda \tilde z_i + (1 - \lambda) \bar z_i)[1:2],
\end{align*}
where a given $\lambda$ determines the unique $z_i[1:2]$. The method for obtaining $\tilde{\mathbf y}_N$ is referred to as the \textbf{search algorithm}, which will be detailed in a later subsection. Building upon the interpolation method, as $\tilde z_i[1:2]$ is a collision-free point, we define $\lambda_{\min} := \{\lambda_{\min,1}, \allowbreak \lambda_{\min,2}, \cdots , \lambda_{\min,N+1}\}$ with
\begin{equation*}
	\lambda_{\min,i} = \max\Big\{ 0,\ \max_{1\le j\le n_{\mathrm{obs}}} \lambda_{ij} \Big\},
\end{equation*}
where, for each obstacle $j$, $\lambda_{ij}$ is the largest $\lambda\in[0,1]$ at which the interpolated position meets the boundary of obstacle $j$, i.e.
\begin{equation*}
	\lambda_{ij} = \max\Big\{ \lambda \in [0,1] \mid \big\| (\lambda \tilde z_i + (1-\lambda)\bar z_i)[1:2] - p_j \big\| = r_j \Big\},
\end{equation*}
with the convention $\lambda_{ij} := 0$ whenever the set on the right-hand side is empty.

$\!\!$Under the definition of $\lambda_{\min,i}$, any $z_i[1:2] = (\lambda \tilde z_i + (1 - \lambda) \bar z_i)[1:2]$ with $\lambda_{\min,i} < \lambda < 1$ is a collision-free  point in $\mathcal M$, while $[\lambda_{\min,i} \tilde z_i + (1- \lambda_{\min,i}) \bar z_i][1:2]$ lies on the edge of one obstacle if $\lambda_{\min,i} > 0$. By further introducing variables $\lambda_{f,i}$ and $\lambda_{d,i}$ with $i = 1, 2, \cdots, N + 1$ and letting
\begin{align}
	z_i &= (1 - \lambda _{\min,i} - \lambda_{f,i} + \lambda_{\min,i}\lambda_{f,i}) \bar{z}_{i} \nonumber  \\
	& \quad +(\lambda_{\min,i} + \lambda_{f,i} - \lambda_{\min,i}\lambda_{f,i}-\lambda_{d,i})\tilde z_i  \nonumber \\ 
	& \quad + \lambda_{d,i} \tilde z_{\max\{1,i-1\}}, \label{z_transfer}
\end{align}
some convex sets of $\mathcal M$ can be extracted as
\begin{align} 
	\mathcal Z_i = \big\{ z_i \big|\eqref{z_transfer}, 0 \leq \lambda_{d,i} \leq \lambda_{f,i} \leq 1 \big\}, 1 \leq i \leq N + 1.
\end{align}
\begin{figure}
	\includegraphics[width=1\linewidth]{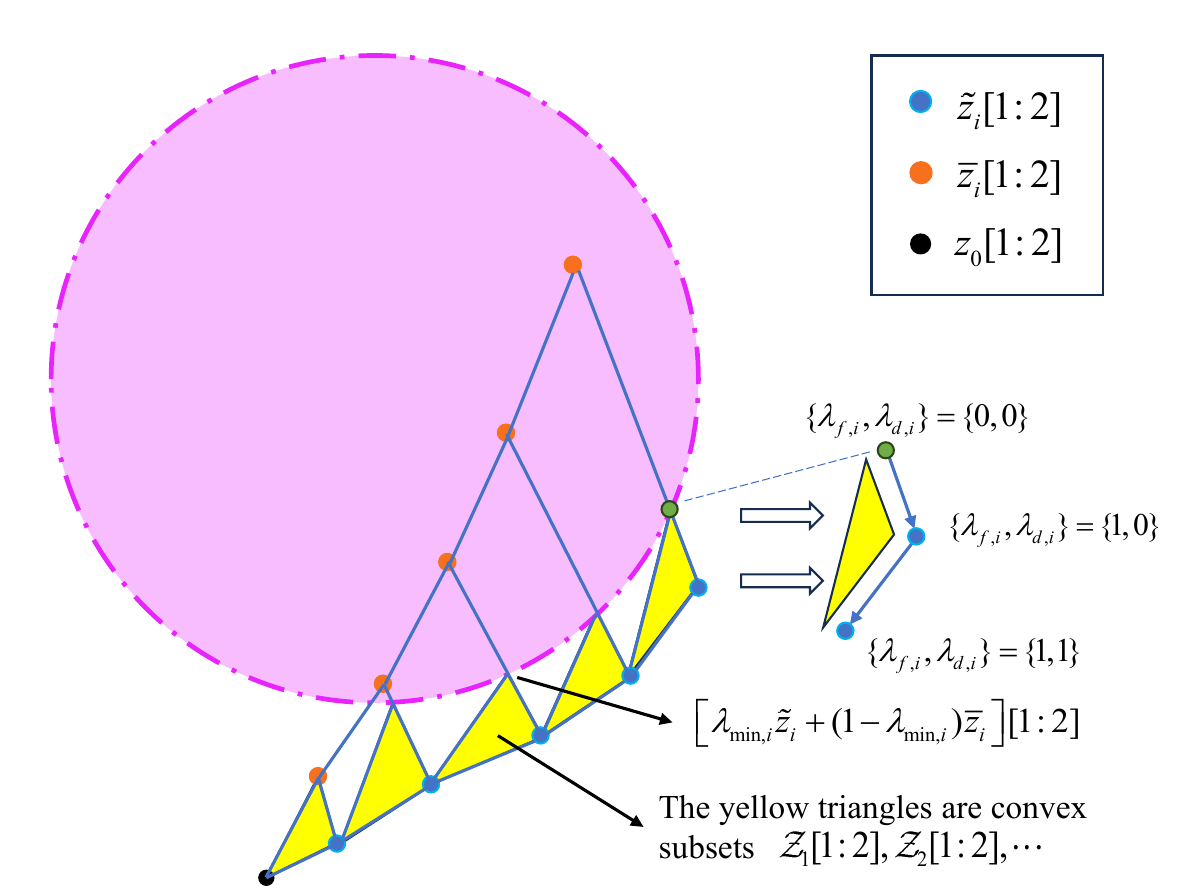}
	\caption{A Visualization of obtaining $\mathcal Z_i$}
	\label{subset}
\end{figure} 
$\!\!$Fig.~\ref{subset} above illustrates the construction and geometric interpretation of $\mathcal Z_i[1:2]$, $\lambda_{\min,i}, \lambda_{f,i}$ and $\lambda_{d,i}$. 

Although the convexity of these extracted sets are guaranteed, the non-feasibility may remain. To further ensure the feasibility, linear inequality constraints on $\lambda_{f,i} , \lambda_{d,i}$ are introduced as
\begin{equation}
	\lambda_{d,i} \leq k_i \lambda_{f,i},  \label{constraint_di}
\end{equation}
where $0 \leq k_i \leq 1$ is pre-selected to keep the convexified area out of obstacles. Define
\begin{align*}
	\lambda_{f} &:= (\lambda_{f,1},\lambda_{f,2},\cdots,\lambda_{f,N+1})^{\top}, \\
	\lambda_{d} &:= (\lambda_{d,1},\lambda_{d,2},\cdots,\lambda_{d,N+1})^{\top},
\end{align*}
and let
\begin{align}
	J_s(\lambda_f, \lambda_d; \rho) &:= J(\mathbf y_N) + \rho \cdot \|z'_0 - z_f\| \|\lambda_f - \mathbf{1}\|^2 \label{Js}
\end{align}
with the weighting parameter $\rho \geq 0$, then the corresponding strict problem can be formulated as $\mathcal P_s(y_0; \tilde{\mathbf y}_N;\bar{\mathbf y}_N;\allowbreak A(y_0), B(y_0))$:
\begin{align}
	& \mathop {\min }\limits_{\{ {\lambda _f}, {\lambda _{d}}\} } \hspace{-7em} &&\hspace{-3em}J_s(\lambda_f, \lambda_d; \rho) \nonumber \\
	& \mathrm{\quad \ \ s. t.} &&\hspace{-3em}0 \leq \lambda_{f,i} \leq 1,   \label{constraint_fi}\\
	&   &&\hspace{-3em}\lambda_{d,i} \geq 0,\\
	&  &&\hspace{-3em}\eqref{z_transfer}, \eqref{DQ_1}, \eqref{DQ_2}, \eqref{c2d_var}, \eqref{new_init}, \eqref{constraint_di}. \nonumber
\end{align} 
According to \eqref{z_transfer} and the given dynamics, the right-hand side of \eqref{Js} is also a function of $\lambda_f, \lambda_d$.

For given $\mathcal P_1'(y_0; N; A(y_0), B(y_0))$, we now discuss the solvability of the corresponding $\mathcal P_r(\cdot), \mathcal P_s(\cdot)$. Clearly, if $\mathcal P_1'(y_0; N; A(y_0), B(y_0))$ is solvable, then $\mathcal P_r(y_0; N; A(y_0), B(y_0))$ is also solvable since it has a relaxed constraint set which is compact. On the other hand, $\mathcal P_s(y_0; \tilde{\mathbf y}_N;\bar{\mathbf y}_N; A(y_0), B(y_0))$ is solvable if $\tilde{\mathbf y}_N$ satisfies the constraints of $\mathcal P_1'(\cdot)$ --- that is, when $\lambda_f = \mathbf 1$ and $\lambda_d = \mathbf 0$. However, if $\mathcal P_1'(y_0; N; A(y_0), B(y_0))$ is unsolvable, specific challenges may occur, which we refer to as connective infeasibility as formalized in Definition 1.
\begin{definition}
	$\mathcal P_1'(y_0; N; A(y_0), B(y_0))$ is said to exhibit \textbf{connective infeasibility} when, despite the problem $\mathcal{P}_0$ being solvable, either the relaxed or strict problem derived from $\mathcal P_1'(y_0; N; A(y_0), B(y_0))$ becomes unsolvable.
\end{definition}  

Connective infeasibility may occur when the terminal state $z_{N+1}$ simultaneously satisfies: (1) the positional proximity to obstacles ($z_{N+1}[1:2]$ near obstacle boundaries), and (2) the conflicting dynamics (e.g. significant velocity toward obstacles). To mitigate the connective infeasibility, the term $\rho$ is introduced in $\mathcal P_s(\cdot)$ to enhance the robustness, where it is referred to as the safety term. Based on the preceding developments, our solution framework can be summarized in Fig.~\ref{fig:subprocess} below.
\begin{figure*}
	\centering
	\includegraphics[width=0.64\linewidth]{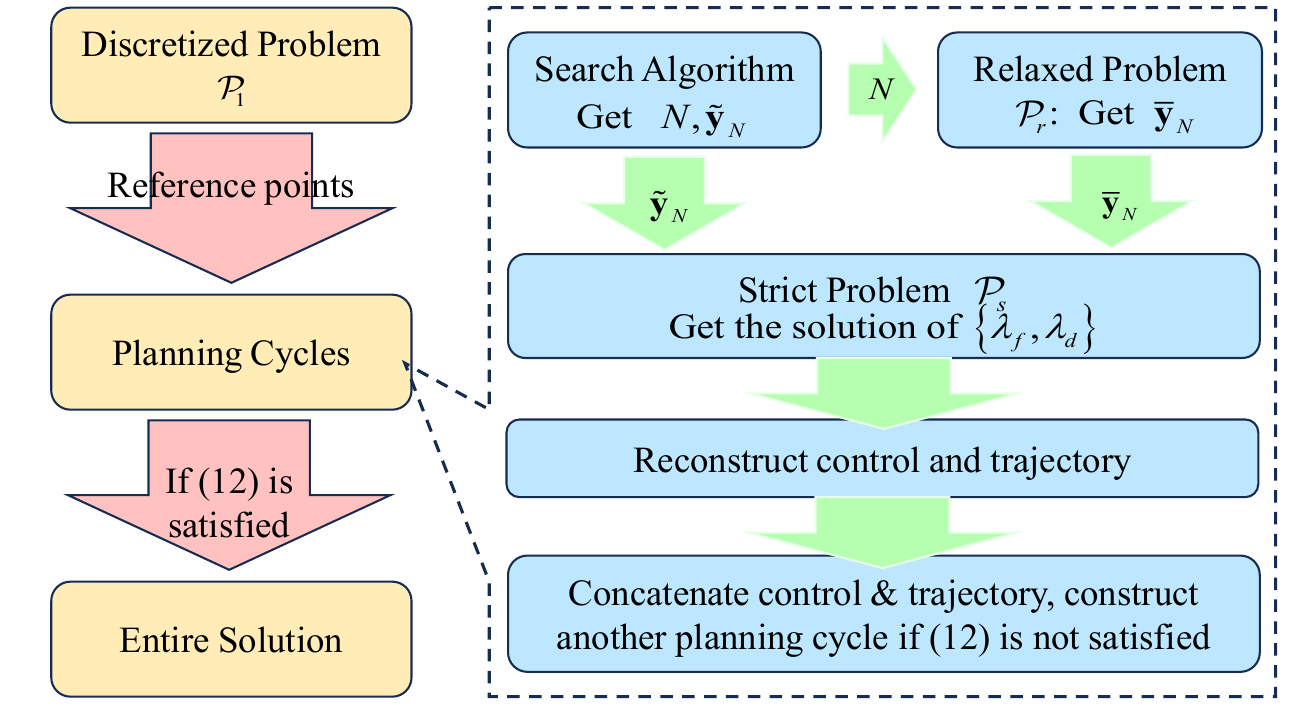}
	\caption{Processes solving $\mathcal P_0$ in the proposed framework}
	\label{fig:subprocess}
\end{figure*}

\subsection{Complete Algorithm}
The entire algorithm of our proposed method is represented as Algorithm 1.
\begin{algorithm}[h] 
	\caption{Online Programming Algorithm}\label{alg:one}
	\begin{algorithmic}[1]
	\STATE{Input}: The map information $\mathcal M$, initial state $z_0$. \\
	\STATE{Output}: A solution of $\mathcal P_0 : \mathcal Y$.
	\STATE $K \gets 1$, $z_0^{(1)} = z_0$, $u_0^{(1)} = 0$, $\mathcal Y \gets \emptyset$, $J^{(0)} = \infty$; \\
	\WHILE{$K < 2$ or $J^{(K-1)} > \gamma^2$}
	{
		\STATE{Take $(z_0^{(K)}, u_0^{(K)}) \to y_0^{(K)}$ as the reference point;}\;
		\STATE{Get $A^{(K)} , B^{(K)} , w^{(K)}$;}
		\STATE{Get a } $\tilde{\mathbf y}^{(K)} = \{\tilde z_1^{(K)}, \cdots ,\tilde z_{N^{(K)}+1}^{(K)}; \tilde u_1^{(K)}, \cdots, \allowbreak \tilde u_{N^{(K)}}^{(K)}\}$ through Search Algorithm;
		\STATE{Construct the relaxed QCQP Problem $\mathcal P^{(K)}_r( y_0^{(K)}; N^{(K)} \allowbreak ; A^{(K)}, B^{(K)}) $;}\;
		\STATE{Get the solution of $\mathcal P^{(K)}_r(\cdot)\Rightarrow \bar{\mathbf y}^{(K)}$;}\;
		\STATE{Solve Problem $\mathcal P^{(K)}_s(y_0^{(K)}; \tilde{\mathbf y}^{(K)}; \bar{\mathbf y}^{(K)}; \cdot) \to \{{\lambda_{f}^{(K)}}, {\lambda_{d}^{(K)}}\}$;}\;
		\STATE Calculate $\mathbf y^{(K)} := \{z_1^{(K)}, \cdots, z_{N^{(K)} + 1}^{(K)}; u_1^{(K)}, \cdots, \allowbreak u_{N^{(K)}}^{(K)}\}$ according to $\tilde{\mathbf y}^{(K)}, \bar{\mathbf y}^{(K)}, \{{\lambda_{f}^{(K)}}, {\lambda_{d}^{(K)}}\}$;
		\STATE Substitute $z_1^{(K)}$ by $z_0^{(K)}$, $z_i^{(K)}$ by $f(z_{i-1}^{(K)}, u_{i-1}^{(K)})$, where $2 \leq i \leq N^{(K)} + 1$ to mitigate error;
		\STATE {$J^{(K)} = \|z_{N^{(K)}+1}^{(K)} - z_f\|_{Q_0}^2$;}
		\IF{$J^{(K)} > \gamma^2$}
			\STATE{Adjust $ N^{(K)}$ to $\hat N^{(K)}$ with $\hat N^{(K)} <N^{(K)} - 1$ and $\hat N^{(K)} \leq N_{\max}$;}\;
		\ELSE
		\STATE{$\hat N^{(K)} \gets N^{(K)}$;}
		\ENDIF
		\STATE{Concatenate} $\mathcal Y$ by $\{ u_1^{(K)},\cdots,u_{\hat N^{(K)}}^{(K)}\}$;
		\STATE Update $(z_0^{(K+1)}, u_0^{(K+1)}) \gets (z_{\hat N^{(K)}+1}^{(K)}, u_{\hat N^{(K)}}^{(K)})$;
		\STATE$K \gets K+1$;\\ 
		\STATE{Update $\rho^{(K)}$;}\;
	}
	\ENDWHILE
	\end{algorithmic}
\end{algorithm} 
In Algorithm 1 and subsequent analysis, we use superscript $(\cdot)^{(K)}$ to represent problems, relative variables and parameters in the $K^{\mathrm{th}}$ planning cycle. To be detailed, let $\mathcal P_{(\cdot)}^{(K)}$, $\mathcal P_1'^{(K)}$ be the abbreviations of $\mathcal P_{(\cdot)}(y_0^{(K)}; N^{(K)}; A^{(K)}, B^{(K)})$, $\mathcal P_1'(y_0^{(K)}; N^{(K)}; A^{(K)}, B^{(K)})$. $\bar z_i^{(K)}$ ($\bar u_i^{(K)}$) and $\tilde z_i^{(K)}$ ($\tilde u_i^{(K)}$) are derived from $\mathcal P_r^{(K)}$ and a feasible solution of the $K^{\text{th}}$ planning cycle respectively by search algorithm. $A^{(K)},B^{(K)},w^{(K)}$ denotes the linearized parameters of $f$ at $y_0^{(K)}$. $\{\lambda_f^{(K)}, \lambda_d^{(K)}\}$ is the optimizer of $\mathcal P_s^{(K)}$, while $z_i^{(K)}, u_i^{(K)}$ is the state-input sequence generated by Algorithm 1: the controls $u_i^{(K)}$ are obtained by solving the linearized subproblem $\mathcal P_1'^{(K)}$ (through $\mathcal P_r^{(K)}$ and $\mathcal P_s^{(K)}$), and the states $z_i^{(K)}$ are then re-simulated through the real dynamics \eqref{c2d_original}, which mitigates the linearization error, so the state-input pair is feasible to the true discretized dynamics.
Algorithm 1 follows a concise iterative structure: it solves the current planning cycle, predicts subsequent states using the calculated controls, concatenates the trajectory, updates the references $z'_0, u'_0$ using the resulting solution, and proceeds to construct and solve the next planning cycle.
This iterative process continues until the quit condition \eqref{boundary_condition_d} is reached. In this process, the entire control is obtained by sequentially concatenating $\{u_1^{(i)}, \cdots , u_{\hat N^{(i)}}^{(i)}\}, i = 1, 2, \cdots$. Note $N^{(K)}$ is initially generated from search algorithms correspondingly, and subsequently it is used as the problem size in both $\mathcal P_r^{(K)}$ and $\mathcal P_s^{(K)}$. To maintain the performance of Algorithm 1, the state-input sequence of one planning cycle is applied only partially by choosing a smaller $\hat N^{(K)}$ and applying $\{ u_1,\cdots,u_{\hat N^{(K)}}\}$. Before proceeding with the algorithm analysis, we will first present two customized trajectory search algorithms which demonstrate good performance.
\subsection{Customized Vortex Artificial Potential Field}
\begin{figure}
	\centering
	\includegraphics[width=0.8\linewidth]{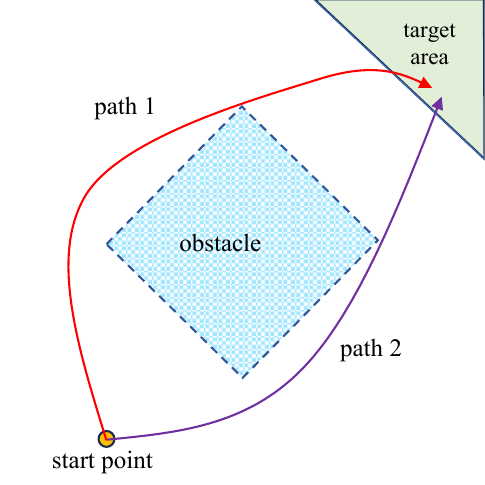}
	\caption{Two possible collision-avoidance paths: sharp turns in the solid red path force the vehicle to decelerate, increasing travel time relative to the smoother dashed blue path.}
	\label{fig:fig2}
\end{figure}

 In this subsection, we consider $z \in \mathbb R^4$ where $z[1:2]$ is for 2-D position and $z[3:4]$ for 2-D velocity. Based on these settings, we propose the Customized Vortex Artificial Potential Field (CVAPF) method as a trajectory search algorithm. Although this algorithm is tailored for double-integrator systems, it readily extends to general affine models, which will be discussed later. Conventional APF normally simulates a virtual force field to navigate while avoiding obstacles, inspired by physical systems under conservative forces. Its enhancement, the Vortex-APF (VAPF) \cite{VAPF_Intro}, extends traditional potential fields by incorporating a vortex field that modulates repulsion into rotational patterns, aiding navigation in complex environments. As in both APF and VAPF, the attractive potential is defined as
\begin{equation}
	P_{att}(z[1:2]) =  \big\|z[1:2] -  z_f[1:2]\big\|^2.
\end{equation}
The repulsive potential acts to drive the vehicle away from obstacles, and we formulate it with the form similar to \cite{APF_REP} as
\begin{equation}
	P_{rep}(z[1:2]) = \xi_{\mathrm{rep}} \sum\limits_{j = 1}^{n_{obs}} {\exp \left( {\frac{{ - \beta  \cdot (\left\| {z[1:2] - {p_j}} \right\| - {r_j})}}{{{r_j}}}} \right)},
\end{equation}
where $\beta$ and $\xi_{\mathrm{rep}}$ are pre-selected positive weights. 
For the APF algorithm, the moving direction can be obtained instantly by calculating $-(\nabla P_{att} + \nabla P_{rep})$ directly. For the VAPF algorithm, the vortex field \(E_v(z[1:2])\) is orthogonal to \(\nabla P_{rep}(z[1:2])\). Its direction \(\mathbf{d}(z[1:2])\) is selected from the unit set
\[
\mathcal{D}(z[1:2]) := \left\{ \tilde{\mathbf{d}} \in \mathrm{Null}(\nabla P_{rep}(z[1:2])) \mid \| \tilde{\mathbf{d}} \| = 1 \right\}.
\]
The field magnitude is scaled by \(\xi_v\), giving \(E_v(z[1:2]) = \xi_v \, \mathbf{d}(z[1:2])\).
In a 2D map, the set \(\mathcal{D}(\cdot)\) contains at least two candidate directions. When \(\nabla P_{rep}(z[1:2]) = \mathbf{0}\), the direction \(\mathbf{d}(z[1:2])\) may be chosen as any unit vector. By incorporating additional directional selection criteria \cite{SAPF}---rather than restricting to a single turning direction as in \cite{LTAPF}—the VAPF algorithm can achieve improved path quality. For instance, it could help to find a shorter way when avoiding an obstacle in the forward of the vehicle, as shown in Fig.~\ref{fig:fig2}. Denote the expected subsequent moving direction of point $\tilde z_i$ as $E_{\tilde z_i}^\dagger$, then the VAPF algorithm calculates
\begin{align}
	{E({{\tilde z}_i}[1:2])}  &=  - \big(\nabla {P_{att}}({\tilde z_i}[1:2]) + \nabla {P_{rep}}({\tilde z_i}[1:2]) \big) \nonumber\\
	& \quad + {E_v}({\tilde z_i}[1:2]), \nonumber
\end{align}
and
\begin{equation}
	E^{\dagger}({\tilde z}_i[1:2])  = \frac{E(\tilde z_i[1:2])}{\|E(\tilde z_i[1:2])\|}. \nonumber
\end{equation}
The vortex field mitigates local minima issues common in APF \cite{OSC}, \cite{APF_LOCAL_MINIMA}, enabling smoother navigation in complex environments.

To further ensure dynamic feasibility, CVAPF introduces a speed search step: once a direction is selected, feasible speeds consistent with physical constraints are computed. If available, a velocity combining direction and admissible speed is chosen; otherwise, a safe deceleration is applied to slow the vehicle while steering it toward the target direction. The complete CVAPF procedure is outlined in Algorithm 2.
\begin{algorithm}[h]
	\caption{Customized VAPF Search Algorithm}\label{alg:two}
\begin{algorithmic}[1]
	\STATE $N \gets 1$,
	 $\tilde z_1 \gets z_0^{(K)},  \mathtt{Not\_Safe} = 1$;\\
	\WHILE{(($N < N_{\min} \ \mathtt{or}$ $\ \mathtt{Not\_Safe}$) $\mathtt{and} \ N \leq N_{\max}$)}
	{
		\STATE{$\tilde z_{N+1}[1:2] = \tilde z_{N}[1:2] + \tilde z_{N}[3:4] \cdot h$;} \\
		\STATE{Calculate $E(\tilde z_{N+1}[1:2]) , E^\dagger(\tilde z_{N+1}[1:2])$;}\\
		\STATE{Search a speed $\varsigma_{N+1}$ from $v_{\max} \to 0$ which meets the constraints with the moving direction;}\\
		\IF{Search is successful}
		{\STATE{ $ \tilde z_{N+1}[3:4] = \varsigma_{N+1} E^\dagger(\tilde z_{N+1}[1:2])$;}}
		\ELSE
		{
			\STATE {$a_{safe} = [a_1 E^\dagger(\tilde z_{N+1}[1:2]) - a_0 \frac{\tilde z_N[3:4]}{\|\tilde z_N[3:4]\|}]$;} \\
			\STATE{$\tilde z_{N+1}[3:4] = \tilde z_{N}[3:4] +  a_{safe} \cdot h$;}\\
		}
		\ENDIF
		\STATE{Update $\xi_v$;}\\
		\STATE{Update $\mathtt{Not\_Safe}$;}\\
		$N \gets N+1$;

	}
	\ENDWHILE
	\IF{$\mathtt{Not\_Safe} = 1$}
	{ \STATE{Algorithm Failed}; }
	\ELSE
	{
		\STATE{Collect $\{\tilde u_1, \tilde u_2, \cdots \}$;}\\
		\STATE{Return Feasible Trajectory $\tilde{\mathbf y}^{(K)}$;}\\
		\STATE{Return $N^{(K)} \gets N-1$;}
	}
	\ENDIF
\end{algorithmic}
\end{algorithm}
In Algorithm 2, $a_{safe}$ is utilized to lower the speed of the agent and alter its direction. For general nonlinear systems, $\varsigma_i$ can be defined by $\|z_i[n_z/2 + 1: n_z]\|$. Since
\begin{equation*}
	\begin{aligned}
		\big\| {{a_{safe}}} \big\| &= \left\| {{a_1}{E^{\dagger}}({\tilde z_{i + 1}}[1:2]) - {a_0}\frac{{{\tilde z_i}[3:4]}}{{\|{\tilde z_i}[3:4]\|}}} \right\| \hfill \\
		&\leq \bigg\| {{a_1}{E^{\dagger}}({\tilde z_{i + 1}}[1:2])} \bigg\| + \left\| {{a_0}\frac{{{\tilde z_i}[3:4]}}{{\|{\tilde z_i}[3:4]\|}}} \right\| \hfill \\
		&= {a_1} + {a_0}, \hfill \\
\end{aligned}
\end{equation*}
$\|a_{safe}\| \leq a_{\max}$ can be easily guaranteed by selecting $a_1$ and $a_0$ with $a_0 + a_1 \leq a_{\max}$. In practical use, a regularization factor can be added to the term $\|\tilde z[3:4]\|$ to avoid singularities. When the vehicle is stationary, i.e., $\tilde z_N[3:4] = \mathbf{0}$, the direction term $\tilde z_N[3:4]/\|\tilde z_N[3:4]\|$ is undefined; in this case the term is omitted, so that the safety acceleration reduces to $a_{safe} = a_1 E^\dagger(\tilde z_{N+1}[1:2])$, which steers the vehicle from standstill toward the desired direction.

The $N$ determined by Algorithm 2 is critical in generating high-quality trajectories, as larger values of $N$ correspond to longer periods and further compromise both real-time performance and solution accuracy. Moreover, we have found that having the trajectory search algorithm terminate at a safer position is helpful to enhance the robustness, for instance, the connective infeasibility can greatly be mitigated. On this basis, Algorithm 2 introduces a $\mathtt{Not\_Safe}$ sign. For a preselected time window $t_w$, define
\begin{equation}
	{T_s} := \Big\{ t \in \left[ {0,{t_w}} \right]\Big| {{\tilde z_{N+1}}[1:2] + {\tilde z_{N+1}}[3:4]  t \in \bigcup\limits_{i = 1}^{n_{obs}} {{\mathcal B_i}} }  \Big\}, \nonumber
\end{equation}
then $\mathtt{Not\_Safe}$ is given by
\[\mathtt{Not\_Safe} = \left\{ \begin{aligned}
	&0, \quad &&T_s = \emptyset, \\
	&1,  &&else,
\end{aligned} \right.\]
and $\min \ T_s$ is also a vital parameter used in the updating of $\rho^{(K)}$, where a smaller $\min \ T_s$ requires bigger $\rho^{(K)}$. Since $N^{(K)}$, which is generated in Algorithm 2 can be large, adjusting $N^{(K)}$ is meaningful to maintain the precision. These steps also show the necessity to substitute $N^{(K)}$ with $\hat N^{(K)}$ in Algorithm 1. Therefore, $N_ {\max}$ is further introduced to confine the length of applied control input. Here $N_{\min} \geq 1$ is the minimum admissible horizon: the search iterates at least until $N \geq N_{\min}$, and $N_{\max} \geq N_{\min}$ caps the horizon from above.

Algorithm~2 is easy to generalize to some affine models \eqref{c2d_var} with $z \in \mathbb R^{n_z} $ and $u \in \mathbb R^{n_u}$. Note that
\[A(y_0) = \left[ {\begin{array}{*{20}{c}}
		\mathbf I_{\frac{n_z}{2}}, & h \mathbf I_{\frac{n_z}{2}} \\
		* & *
\end{array}} \right] \in \mathbb R^{n_z \times n_z} \]
where the first $n_z/2$ rows describe the kinematic update of the pose by the generalized velocity, $z_{i+1}[1:n_z/2] = z_i[1:n_z/2] + h\, z_i[n_z/2+1:n_z]$. The remaining rows, together with $B(y_0)$ and $w(y_0)$, encode the model-dependent dynamics. The extension of Algorithm~2 is then immediate: the potential-field computation can be extended to so called state-level, which uses the generalized pose components $\tilde z_i[1:n_z/2]$. The speed search, in turn, depends only on the generalized velocity magnitude $\|\tilde z_i[n_z/2+1:n_z]\|$, as discussed above. In the safe-deceleration step, the velocity direction $z_i[3:4]/\|z_i[3:4]\|$ is replaced by its generalized counterpart $\tilde z_i[n_z/2+1:n_z]/\|\tilde z_i[n_z/2+1:n_z]\|$. Consequently, CVAPF serves as a search algorithm for the entire class of linearized dynamics \eqref{c2d_var}, covering the double-integrator, unicycle, Ackermann, and planar-quadrotor dynamics described in Section~2.

\subsection{Customized Dynamic Window Approach}
The Customized Dynamic Window Approach (CDWA) is designed as another search algorithm for nonlinear models based on dynamic window approach (DWA) \cite{DWA}. By judiciously balancing the trade-offs between speed and safety, DWA enables agents to navigate dynamic environments with minimal risks. 
The DWA centers on the concept of a ``dynamic window.'' This window represents the set of feasible control inputs available to an agent in its current state. Its construction incorporates multiple state factors, including position, orientation, and velocity. System-specific constraints are also integral to this formulation. From the agent's perspective, it is essential to predict proper actions based on its current state---i.e., to construct a dynamic window. 
From the algorithm designer’s standpoint, defining a policy to select actions from these feasible sets is equally critical.

To adapt DWA for our framework, we introduce CDWA. Following the standard DWA procedure, we first compute the dynamic window at the current state by enumerating control inputs consistent with the constraints. A subset of these inputs is then extracted for evaluation based on the map information, along with the $\mathtt{Not\_Safe}$ flag and $\min T_s$. The state at the next time step is predicted using the chosen input, and the current state is updated accordingly. Repeating this process over several steps yields a trajectory of $N$ inputs and $N+1$ states. The pseudocode of CDWA is given in Algorithm 3 below.

\begin{algorithm}[h]
\caption{Customized DWA Search Algorithm }\label{alg:three}
\begin{algorithmic}[1] 
	\STATE{$N \gets 0$, $z_1 \gets z'_0$;} \\
	\WHILE{(($N < N_{\min} \ \mathtt{or}$ $\ \mathtt{Not\_Safe}$) $\mathtt{and} \ N \leq N_{\max}$)}
	{
		\STATE{Detect Obstacles along the direction of current velocity;}
		\STATE{Calculate the dynamic window;}\;
		\STATE{Obtain  $\min \ T_s $;}\;
		\STATE{Obtain current input;}\; 
		\STATE{Update $\mathtt{Not\_Safe}$;}\;
		\STATE{$N \gets N+1$;}\;
	}
	\ENDWHILE
	\STATE{Compute the control sequence and return the feasible trajectory $\tilde{\mathbf y}^{(K)}$ and the horizon $N^{(K)}$, as in the finalization steps of Algorithm 2;}
\end{algorithmic}
\end{algorithm}
\subsection{Theoretical Guarantees}
We subsequently show the convergence property of Algorithm 1 under a mild assumption given in Assumption~\ref{todo20241117}. For the $K$-th planning cycle, let
\begin{equation}
	\tilde J^{(K)} := J(\mathbf y^{(K)}) + \rho^{(K)} \|z_0^{(K)} - z_f\| \|\lambda_f^{(K)} - \mathbf{1}\|^2, \label{tilde_J}
\end{equation}
where $\mathbf y^{(K)}$ is the trajectory re-simulated through the real dynamics \eqref{c2d_original} in Algorithm~1, so that $J(\mathbf y^{(K)}) = \|z_{N^{(K)}+1}^{(K)} - z_f\|_{Q_0}^2$, and $\{\lambda_f^{(K)}, \lambda_d^{(K)}\}$ is an optimal solution of $\mathcal P_s^{(K)}$.
\begin{assumption}
	\label{todo20241117}
	There exists a contraction factor $0 \leq \kappa < 1$, which makes
	\begin{equation}
		\frac{\tilde J^{(K)}}
		{\tilde J^{(K-1)}} \leq \kappa  \label{ass3}
	\end{equation}
	hold for all but finitely many iterations $K$. In other words, the number of iterations that violate \eqref{ass3} is limited. 
\end{assumption} 

We now discuss two complementary cases to justify Assumption~\ref{todo20241117}. In Case I, suppose $\exists \; C > 1$ with $\tilde J^{(K-1)} > C \gamma^2$. The generalized velocity-related cost in $\tilde J$ can be made relatively small by a suitable choice of $Q_0$, so that $\tilde J$ is dominated by the position error. With the assistance of the search algorithm, the inequality $\tilde J^{(K)} \leq \tilde J^{(K-1)}$ is then naturally satisfied, provided the initial generalized velocity $z^{(K)}_{0}[n_z/2 + 1: n_z]$ remains moderate. If the initial velocity is instead undesirable, e.g., the vehicle moves away from the target region at maximum speed, the contraction bound \eqref{ass3} is still satisfied after several cycles. Moreover, when obstacles are encountered, a strategic choice of the maneuver direction is required. Since each obstacle is approximated by a closed ball, there always exists at least one direction that moves the vehicle closer to the target region, and the search algorithm can prioritize such a direction. Case II is complementary to Case I: when $\tilde J^{(K-1)} > C \gamma^2$ for some $C > 1$ does not hold, a proper update of the safety term $\rho^{(K)}$ can drive the vehicle to satisfy the boundary constraint; the convexity of $\mathcal P_s$ ensures that this is feasible whenever the search algorithm can divert the vehicle toward the target region. Thus, Assumption~\ref{todo20241117} is mild. Together with Assumption~1, it guarantees that the target region is reachable within finitely many planning cycles, as established in Theorem~\ref{theorem_1}.
\begin{theorem}\label{theorem_1}
	Under Assumptions~\ref{original_A1} and \ref{todo20241117}, Algorithm~1 terminates after finitely many planning cycles. Specifically, there exists $K < \infty$ such that the terminal state of the $K$-th cycle satisfies $\| z_{N^{(K)}+1}^{(K)} - z_f \|_{Q_0} \leq \gamma$.
\end{theorem}
\textbf{\textit{Proof.}} Let $z_{N^{(K)}+1}^{(K)}$ denote the terminal state of the $K$-th planning cycle, i.e., the terminal state of the re-simulated trajectory $\mathbf y^{(K)}$ in \eqref{tilde_J}. Since the safety term in \eqref{tilde_J} is nonnegative and $J(\mathbf y^{(K)}) = \|z_{N^{(K)}+1}^{(K)} - z_f\|_{Q_0}^2$ by \eqref{J_P_1}, we have $\tilde J^{(K)} \geq \|z_{N^{(K)}+1}^{(K)} - z_f\|_{Q_0}^2$; hence the quit condition \eqref{boundary_condition_d} holds as soon as $\tilde J^{(K)} \leq \gamma^2$. The case $\kappa = 0$ is immediate, for then \eqref{ass3} forces $\tilde J^{(K)} = 0$ for all but finitely many $K$, so that the quit condition is met; we henceforth assume $0 < \kappa < 1$.

Under Assumption~\ref{todo20241117}, the ratio $R_K := \tilde J^{(K)}/\tilde J^{(K-1)}$ satisfies $R_K \leq \kappa$ for all but finitely many iterations $K$. Suppose, for contradiction, that Algorithm~1 does not terminate after finitely many planning cycles. Then every cycle is non-terminal, so that $\tilde J^{(K)} \geq \|z_{N^{(K)}+1}^{(K)} - z_f\|_{Q_0}^2 \geq \gamma^2 > 0$ for all $K$, and all ratios below are well defined. Let $K_1 < \cdots < K_m$ be the finitely many iterations at which $R_K > \kappa$, and set $\kappa_i := R_{K_i}$, $i = 1, \cdots, m$; each $\kappa_i$ is a finite positive constant. For any $K \geq K_m$, telescoping the ratios $R_2, \cdots, R_{K+1}$ yields
\begin{equation}
	\frac{\tilde J^{(K+1)}}{\tilde J^{(1)}} = \prod\limits_{j = 2}^{K+1} R_j \leq \kappa^{K - m} \prod\limits_{i = 1}^m {\kappa _i}, \nonumber
\end{equation}
since among the $K$ factors exactly the $m$ ratios $R_{K_1}, \cdots, R_{K_m}$ exceed $\kappa$. Combining this bound with $\tilde J^{(K+1)} \geq \|z_{N^{(K+1)}+1}^{(K+1)} - z_f\|_{Q_0}^2$, we obtain
\begin{align}
	\left\| {z_{N^{(K+1)}+1}^{(K+1)} - {z_f}} \right\|_{Q_0}^2 &\leq \tilde J^{(K+1)} \nonumber \\
	& \leq  \tilde J^{(1)} {\kappa ^{K - m}} \prod\limits_{i = 1}^m {{\kappa _i}}. \nonumber
\end{align}
Since $0 < \kappa < 1$, the right-hand side tends to zero as $K \to \infty$. Taking
\begin{equation}
	K > m + \frac{{\log \frac{\gamma^2}{\tilde J^{(1)}} - \sum\limits_{i = 1}^m {\log {\kappa _i}} }}{{\log \kappa }}, \nonumber
\end{equation}
we get $\tilde J^{(K+1)} \leq \gamma^2$ and hence $\| z_{N^{(K+1)}+1}^{(K+1)} - z_f \|_{Q_0} \leq \gamma$; cycle $K+1$ therefore satisfies the quit condition, which contradicts the assumption that Algorithm~1 never terminates. Consequently, Algorithm~1 terminates after finitely many planning cycles. \hfill $\square$

Building on Theorem~\ref{theorem_1}, we now establish the candidate optimal property of the concatenated solution. Two additional assumptions are required, stated as Assumptions~\ref{Full_Rank} and~\ref{KKT}.
\begin{assumption}\label{Full_Rank}
	For each planning cycle $\mathcal{P}_1'(y_0^{(K)}; N^{(K)};\allowbreak A^{(K)}, B^{(K)})$, the matrix $B^{(K)}$ has full column rank. Moreover, there exists an integer $\tau \geq 1$ such that every $\tau$-step reachability matrix
	\begin{equation}
		\mathcal C_i := \left[ B_i, \; A_iB_{i-1}, \; \cdots, \; A_i \cdots A_{i-\tau+2}B_{i-\tau+1} \right] \label{reach_matrix}
	\end{equation}
	formed along an optimizer of a planning cycle has full row rank $n_z$, where $A_i := \partial f/\partial z_i$ and $B_i := \partial f/\partial u_i$ are evaluated at $(z_i, u_i)$ (for the linearized problem $\mathcal P_1'(\cdot)$ they are the matrices $A_i, B_i$ of the affine dynamics).
\end{assumption}
\begin{remark}
	The full‑rank condition in Assumption~\ref{Full_Rank} is standard and satisfied by the vehicle models in Section 2.1; it enables input reconstruction from consecutive states (Appendix B). The $\tau$-step condition imposes reachability on the optimizer‑linearized dynamics, which enforces stationarity at stitching states by forcing transmitted costates to vanish across the stitch (Appendix A). Since $\mathcal C_i$ has $\tau n_u$ columns, $\tau \geq n_z/n_u$ is necessary. Specifically, $\tau=1$ requires $(\partial f/\partial u)^\top$ to be injective and cannot hold for the underactuated models in Section 2.1 with $n_u = n_z/2$, where $\tau=2$ becomes necessary and sufficient. For the double‑integrator model,
	\[
	\mathcal C_i = \big[ B_i,\; A_iB_{i-1} \big] =
	\begin{bmatrix}
		\mathbf O_2 & h^2 \mathbf I_2 \\
		h\mathbf I_2 & h\mathbf I_2
	\end{bmatrix},
	\]
	the condition holds unconditionally for all $h>0$. For the unicycle model, $\tau=2$ suffices when the vehicle is in motion; degeneracy arises only at complete standstill over the time window.
\end{remark}

Now, let $\mathscr L_N(z'_0;\cdot) $ denote the Lagrangian function of $\mathcal P_1(z'_0;N; \cdot)$. Namely, in $\mathcal P_1(z'_0;N)$, we define
\begin{align}
	&\mathscr L_{N}(z'_0;z_i, u_i, \upsilon_i, \zeta_{ij}, s_i, q_i) := \left\| {{z_{N+1 } - z_f}} \right\|_{{Q_0}}^2 \nonumber \\
	&\quad + \sum\limits_{i = 1}^{N} {{\upsilon _i}\left( {{z_{i + 1}} - f(z_i, u_i)} \right)}  \nonumber \\
	& \quad + \sum\limits_{i = 1}^{N + 1} {\sum\limits_{j = 1}^{{n_{obs}}} {{\zeta _{ij}}\left( {r_j^2 - {{\left\| {{z_i}[1:2] - {p_j}} \right\|}^2}} \right)} } \nonumber \\
	& \quad + \sum\limits_{i = 1}^{N+1} {{s_i}{G_1}({z_i})}  + \sum\limits_{i = 1}^{N} {{q_i}{G_2}({u_i})}, \label{lagrange}
\end{align}
where $\upsilon_i, \zeta_{ij}, s_i, q_i$ are Lagrange multipliers.
\begin{assumption}
	\label{KKT}
	The local optimizer of \(\mathcal{P}_1(z'_0; N)\) (if exists) satisfies the Mangasarian--Fromovitz constraint qualification (MFCQ) \cite{MFCQ}. Specifically, at such an optimizer, the gradients of all active equality constraints are linearly independent, and there exists a vector that is:

    1). a descent direction for all active inequality constraints;
    
    2). orthogonal to the gradients of all equality constraints.

\end{assumption}

Under Assumption 5, if $\mathcal P_1(\cdot)$ admits a local optimizer, then there exists a pair of Lagrange multipliers such that the optimizer satisfies the Karush–Kuhn–Tucker (KKT) conditions. This result further leads to Theorem 2.
\begin{theorem}
	\label{revised_th_2}
	Under Assumptions~\ref{Full_Rank} and~\ref{KKT}, with $\tau$ the integer of Assumption~\ref{Full_Rank}, let $\{z_1^1, z_2^1, \cdots, z_{m+1}^1; u_1^1, \cdots, u_m^1\}$ be a local optimizer of $\mathcal{P}_1(z'_0; m)$, and let $\{z_1^2, z_2^2, \cdots, z_{n+1}^2; u_1^2, \cdots, u_n^2\}$ be a local optimizer of $\mathcal{P}_1(z_{m'+1}^1; n)$ for some $m' \leq m-1$ with $m' + \tau - 1 \leq m$ and $\tau \leq n$. If the $\tau$ inputs on either side of the stitching state are strictly feasible for the input bound, i.e.,
	\[
	\begin{cases}
		{u_{i}^1}^{\top}Q_2 u_{i}^1<1, & m' \leq i \leq m'+\tau-1, \\[4pt]
		{u_{i}^2}^{\top}Q_2 u_{i}^2<1, & 1 \leq i \leq \tau,
	\end{cases}
	\]
	and the corresponding states, including the stitching state $z_{m'+1}^1 = z_1^2$, are strictly feasible for the collision and state-bound constraints, i.e., $\|z_i^\ell[1:2] - p_j\| > r_j$ for all $j$ and $G_1(z_i^\ell) < 0$ for $m'+1 \leq i \leq m'+\tau-1$ when $\ell = 1$ and for $1 \leq i \leq \tau$ when $\ell = 2$, then the concatenated sequence
	\begin{align*}
		\big \{z_1^1,\dots,z_{m'+1}^1,z_2^2,\dots,z_{n+1}^2; u_1^1,\dots,u_{m'}^1,u_1^2,\dots,u_n^2\big \}
	\end{align*}
	satisfies the KKT conditions of $\mathcal{P}_1(z'_0; m'+n)$ with the multipliers obtained by concatenating the corresponding multipliers of the two subproblems.
\end{theorem}
\noindent \textbf{\textit{Proof.}} See Appendix A.  \hfill $\square$

Theorem~2 shows that concatenating the optimizers of two adjacent planning cycles preserves the KKT conditions of the combined problem. It does not assert that the concatenated trajectory is itself a local optimizer of $\mathcal P_1(z_0'; m'+n)$; rather, it identifies a set of multipliers for which the concatenated solution satisfies the first-order necessary conditions. While this falls short of a full optimality guarantee, the result is valuable, as it provides a theoretical foundation for the core operation of Algorithm~1---the sequential concatenation of cycle solutions---by confirming that no first-order inconsistency arises at the stitching point. 

For the approximated problem $\mathcal P_1'(\cdot)$, a similar concatenation result holds for linearized dynamics, as stated in Corollary~\ref{col.1}.
\begin{corollary}\label{col.1}
	Under Assumptions~\ref{Full_Rank} and~\ref{KKT}, let $\{z_1^1, \cdots, z_{m+1}^1; u_1^1, \cdots, u_m^1\}$ be a local optimizer of the linearized problem $\mathcal P_1'(y_0; m; \cdot)$ with reference pair $y_0 := (z'_0, u'_0)$ and affine dynamics $z_{i+1} = A_i^1 z_i + B_i^1 u_i + w_i^1$, and let $\{z_1^2, \cdots, z_{n+1}^2; u_1^2, \cdots, u_n^2\}$ be a local optimizer of the linearized problem $\mathcal P_1'(y_0'; n; \cdot)$ with reference pair $y_0' := (z_{m'+1}^1, u_{m'}^1)$ and affine dynamics $z_{i+1} = A_i^2 z_i + B_i^2 u_i + w_i^2$, for some $m' \leq m-1$ with $m' + \tau - 1 \leq m$ and $\tau \leq n$. If the inputs $u_i^1$ ($m' \leq i \leq m'+\tau-1$) and $u_i^2$ ($1 \leq i \leq \tau$) are strictly feasible for the input bound, and the states $z_i^1$ ($m'+1 \leq i \leq m'+\tau-1$) and $z_i^2$ ($1 \leq i \leq \tau$) are strictly feasible for the collision and state-bound constraints, then the concatenated sequence
	\begin{align*}
		\big\{z_1^1, \cdots, z_{m'+1}^1, z_2^2, \cdots, z_{n+1}^2;\, u_1^1, \cdots, u_{m'}^1, u_1^2, \cdots, u_n^2\big\}
	\end{align*}
	satisfies the KKT conditions of the linearized problem $\mathcal P_1'(y_0; m'+n; \cdot)$ with affine dynamics $z_{i+1} = A_i z_i + B_i u_i + w_i$, where $(A_i, B_i, w_i)$ are given by
	\[
	(A_i, B_i, w_i) = \left\{
	\begin{array}{ll}
		(A_i^1, B_i^1, w_i^1), & \quad i \le m', \\
		(A_{i - m'}^2, B_{i - m'}^2, w_{i - m'}^2), & \quad i > m'.
	\end{array}
	\right.
	\]
\end{corollary}

Theorem 2 and Corollary 1 establish the soundness of concatenating segmented solutions and, consequently, justify the receding-horizon adjustment of $N^{(K)}$ in Algorithm 1. Although the analysis has been carried out for two segments, the conclusion extends directly to trajectories assembled over an arbitrary number of planning cycles, thereby justifying the iterative solution of each linearized subproblem $\mathcal P_1'(y_0^{(K)}; N^{(K)}; A^{(K)}, B^{(K)})$ in Algorithm 1. Moreover, when the relaxed solution is collision-free (i.e., $\lambda_{\min,i} = 0$ for all $i$), the convex reformulation $\mathcal P_r/\mathcal P_s$ is lossless: each planning-cycle solution can reach a global optimizer of its own subproblem, not merely a KKT point (see Theorem~4), even if non-convex constraints exist.

In what follows, we turn to another critical index: the time cost of the entire control. Since the discretization is based on fixed step lengths, Algorithm 1 accumulates this cost from the horizons $\hat N^{(K)}$ of its planning cycles. Specifically, the final time of the entire control is given by
${t_f} = h\sum_{i = 1}^{K - 1} \hat N^{(i)} + h\hat N^{(K)}.$
Since $h\hat N^{(K)} \le hN_{\max}$, a smaller $N_{\max}h$ yields a more accurate estimate of $t_f$. However, an exact value of $t_f$ is not required to establish the ``real-time optimality'' of the entire control; rather, it suffices to show that $t_f$ approximately attains the locally minimal final time.
\begin{theorem}\label{thm_time_opt}
	Suppose Algorithm~1 generates iterates $\mathbf y^{(1)},\dots,\mathbf y^{(K)}$ and terminates at cycle $K$ under the stopping condition $\|z_{N^{(K)}+1}^{(K)} - z_f\|_{Q_0}^2 \le \gamma^2$. Let $t_f := h\sum_{i=1}^{K}\hat N^{(i)}$. If the concatenated state‑input sequence $\{z_1^{(1)},\dots,z_{N^{(K)}+1}^{(K)}; u_1^{(1)},\dots,u_{N^{(K)}}^{(K)}\}$ is a local minimizer of $\mathcal P_1\big(z_0^{(1)};\sum_{i=1}^{K}\hat N^{(i)}\big)$, then for all feasible perturbed trajectories of $\mathcal P_1(\cdot)$ in a sufficiently small neighborhood: either the target region is attained within cycle $K$, giving $t_f(\delta)=t_f$; or at least one extra planning cycle is needed, yielding $t_f(\delta)-t_f\ge h$, where $t_f(\delta)$ is the terminal time of the perturbed solution for $\mathcal P_0$. Furthermore, under Assumption~\ref{todo20241117}, $\|z_{N^{(K)}+1}^{(K)} - z_f\|_{Q_0} \to \gamma^-$ as $N_{\max}h\to 0$, so $t_f$ converges to a locally minimal arrival time.
\end{theorem}
\textbf{\textit{Proof.}} Let $\{z_1^{(1)}(\delta), \dots, z_{N^{(K)}+1}^{(K)}(\delta); u_1^{(1)}(\delta), \dots, u_{N^{(K)}}^{(K)}(\delta)\}$ be any trajectory feasible for $\mathcal P_1(\cdot)$ in a sufficiently small neighborhood of the concatenated solution, and let $t_f(\delta)$ denote the final time of the corresponding control when applied to $\mathcal P_0$. Since the concatenated sequence is a local optimizer of $\mathcal P_1(\cdot)$, we further have
\[\|z_{N^{(K)}+1}^{(K)}(\delta) - z_f\|_{Q_0}^2 \geq \|z_{N^{(K)}+1}^{(K)}- z_f\|_{Q_0}^2.\]
If the perturbed trajectory additionally satisfies $(z_{N^{(K)}+1}^{(K)}(\delta) - z_f)^\top Q_0 (z_{N^{(K)}+1}^{(K)}(\delta) - z_f) > \gamma^2$, then the quit condition is not met at the $K$-th cycle and Algorithm~1 continues with at least one additional planning cycle; for the corresponding control $U(\delta)$ of $\mathcal P_0$, it follows that
$t_f(\delta) - t_f \geq h$.
Otherwise, if $(z_{N^{(K)}+1}^{(K)}(\delta) - z_f)^\top Q_0 (z_{N^{(K)}+1}^{(K)}(\delta) - z_f) \leq \gamma^2$, the target region is reached within the horizon of the $K$-th cycle and $t_f(\delta) = t_f$. This establishes the first claim. It remains to show that $t_f$ approaches the locally minimal final time as $N_{\max}h \to 0$. To this end, rewrite the quit condition in the equivalent form
\begin{equation}
	\|z_{N^{(K)}+1}^{(K)} - z_f\|^2_{Q_0} \leq \gamma^2.
\end{equation}
Since the $(K-1)$-th cycle does not terminate, $\|z_{N^{(K-1)}+1}^{(K-1)} - z_f\|_{Q_0} > \gamma$, whereas the $K$-th cycle satisfies $\|z_{N^{(K)}+1}^{(K)} - z_f\|_{Q_0} \leq \gamma$. By the boundedness of states, there must exist an upper bound of $\{\|z_{N^{(K)}+1}^{(K)} - z_f\|\}_K$, denoted as $C_{\max}$. To quantify the gap between the two terminal costs, we have
\begin{align}
	& \left| \|z_{N^{(K)}+1}^{(K)} - z_f\|^2_{Q_0} -	\|z_{N^{(K-1)}+1}^{(K-1)} - z_f\|^2_{Q_0} \right| \label{time_opt_2} \\
	& \leq 2C_{\max} \|z_{N^{(K)}+1}^{(K)} - z_{N^{(K-1)}+1}^{(K-1)}\| \|Q_0\| \label{11} \\
	& \leq 2 C_{\max} \cdot  \Bigg ( \left\| {z_{{N^{(K)}} + 1}^{(K)} - z_{{{\hat N}^{(K - 1)}} + 1}^{(K - 1)}} \right\| \nonumber \\
	& \quad + \left\| {z_{{{\hat N}^{(K - 1)}} + 1}^{(K - 1)} - z_{{N^{(K - 1)}} + 1}^{(K - 1)}} \right\| \Bigg) \cdot \|Q_0\|.   \label{time_opt_1}
\end{align}
Note that the discrete dynamics \eqref{c2d_original} are obtained by forward Euler discretization of \eqref{non_linear_original}, so the one-step state increment satisfies $\Delta_i^{(K)} := \|z_{i+1}^{(K)} - z_i^{(K)}\| = h\,\|f_c(z_i^{(K)},u_i^{(K)})\|$ for any (linear or nonlinear) $f_c$. Since the state and input sequences are bounded and $f_c$ is differentiable, there exists a constant $M_f$ such that $\|f_c(z_i^{(K)},u_i^{(K)})\| \le M_f$, giving $\Delta_i^{(K)} \le hM_f$.

For the first norm on the right-hand side of \eqref{time_opt_1}, we further obtain
\begin{align}
	\left\| {z_{{N^{(K)}} + 1}^{(K)} - z_{{{\hat N}^{(K - 1)}} + 1}^{(K - 1)}} \right\| 
	& = \left\| {z_{{N^{(K)}} + 1}^{(K)} - z_1^{(K)}} \right\| \nonumber \\
	& \leq  \sum\limits_{i = 1}^{N^{(K)}} \underbrace{{\left\| {z_{i+1}^{(K)} - z_i^{(K)}} \right\|}}_{\Delta_i^{(K)}} \nonumber \\
	& \leq h \cdot N_{\max} M_f. \label{bounded}
\end{align}
Similarly,
\begin{align}
	{\left\| {z_{{{\hat N}^{(K - 1)}} + 1}^{(K - 1)} - z_{{N^{(K - 1)}} + 1}^{(K - 1)}} \right\|} 
	&\le \sum_{i = {{\hat N}^{(K - 1)}} + 1}^{{N^{(K-1)}}} \Delta_i^{(K-1)} \nonumber \\
	&\leq h \cdot N^{(K -1)} M_f.  \label{bounded_2}
\end{align}
\eqref{bounded} and \eqref{bounded_2} give $N_{\max}\max_i\Delta_i^{(K)} \to 0$ and $N_{\max}\max_i\Delta_i^{(K-1)} \to 0$ as $N_{\max}h \to 0$, so \eqref{time_opt_1} implies $\eqref{time_opt_2} \to 0$ as $N_{\max}h \to 0$. Combining this with $\| z_{{N^{(K-1)}} + 1}^{(K-1)} - z_f \|_{Q_0} > \gamma$ and $\| z_{{N^{(K)}} + 1}^{(K)} - z_f \|_{Q_0} \leq \gamma$, we obtain $\| z_{{N^{(K)}} + 1}^{(K)} - z_f \|_{Q_0} \to \gamma^-$ as $N_{\max}h \to 0$. Hence, for a sufficiently small $N_{\max}h$, every perturbed control $\{u^{(1)}_1(\delta),\cdots,u^{(K)}_{N^{(K)}}(\delta)\}$ fails to satisfy the quit condition except the concatenated solution itself. Consequently, $t_f = h\sum\nolimits_{i = 1}^K {{{\hat N}^{(i)}}}$ converges to a locally minimal final time. This completes the proof. \hfill $\square$

Although $t_f$ converges to a locally minimal final time as $N_{\max}h \to 0$ under Theorem~3's precondition, an excessively small $N_{\max}h$ may give rise to undesirable phenomena, e.g., the conditions of Theorem~\ref{revised_th_2} may become difficult to satisfy. A lower bound on $N_{\max}h$ should therefore be imposed in practice, and $t_f$ only approximates the locally minimal control time.

Next, we provide sufficient conditions under which the optimal solution of $\mathcal P_s^{(K)}$ certifies the KKT conditions --- i.e., a candidate local optimizer --- of $\mathcal P_1'(y_0^{(K)};N^{(K)};A^{(K)},B^{(K)})$ over the planning-cycle region. We begin with defining the relative interior solution.
\begin{definition}
	A feasible solution to $\mathcal{P}_s^{(K)}(\cdot)$ is called a \emph{relative interior solution} if it satisfies every inequality constraints of $\mathcal{P}_s^{(K)}(\cdot)$ with strict inequality: for all $1 \leq i \leq N^{(K)} + 1$, $0 < \lambda_{d,i}^{(K)} < \lambda_{f,i}^{(K)} < 1$ and $\lambda_{d,i}^{(K)} < k_i^{(K)} \lambda_{f,i}^{(K)}$, and the state--input bounds are strict, $G_1(z_i^{(K)}) < 0$ for $1 \le i \le N^{(K)}+1$ and $G_2(u_i^{(K)}) < 0$ for $1 \le i \le N^{(K)}$, with $z_i^{(K)}$ given by \eqref{z_transfer}. \label{def_rel_int}
\end{definition}
\begin{theorem}\label{thm:4}
	Under Assumptions~\ref{Full_Rank} and~\ref{KKT}, assume $\tilde z_i^{(K)} - \bar z_i^{(K)}$ and $\tilde z_i^{(K)} - \tilde z_{i-1}^{(K)}$ of the interpolation \eqref{z_transfer} are nonparallel for each $2 \leq i \leq N^{(K)}+1$, so that \eqref{z_transfer} parameterizes the planning-cycle region $\mathcal R^{(K)} := \prod_{i = 1}^{N^{(K)}+1}\mathcal Z_i^{(K)}$ bijectively. Let $\{\lambda_f^{(K)}, \lambda_d^{(K)}\}$ be an optimal solution of $\mathcal P_s^{(K)}$ with $\rho^{(K)} = 0$, and let $\{z_1^{(K)}, \cdots, z_{N^{(K)}+1}^{(K)}; u_1^{(K)}, \cdots, u_{N^{(K)}}^{(K)}\}$ be the state-input sequence generated via \eqref{z_transfer} and the affine dynamics \eqref{c2d_var}. Then this is a candidate local optimizer of $\mathcal P_1'(y_0^{(K)}; N^{(K)}; A^{(K)}, B^{(K)})$ over the region $\mathcal R^{(K)}$, if either of the following holds:	 
	
		1). $\lambda_{\min,i}^{(K)} = 0$ for all $i = 1, \dots, N^{(K)}+1$, or $\tilde z_i^{(K)} = \bar z_i^{(K)}$ for all $i = 1, \dots, N^{(K)}+1$;
	
		2). $\{\lambda_f^{(K)}, \lambda_d^{(K)}\}$ is a relative interior solution of $\mathcal P_s^{(K)}$.
\end{theorem}
\textbf{\textit{Proof.}} See Appendix B. \hfill $\square$
\begin{remark}
	Theorem~\ref{thm:4} imposes two preconditions. The first holds whenever the solution of $\mathcal{P}_r(\cdot)$ is collision‑free; the second is satisfied if the optimizer of $\mathcal{P}_1'(\cdot)$ lies strictly inside the constructed convex region. The first condition generally holds over map regions without excessive obstacle density. The nonparallel‑anchors hypothesis is generic: it fails only when two successive search states and the associated relaxed state are collinear, and it renders the stationarity transfer in Appendix~B exact.
\end{remark}

We have analyzed properties of Algorithm 1 under the assumption that the safety term is disabled, i.e., $\rho^{(j)} = 0$ in each $\mathcal P_s^{(j)}$. We now briefly examine the effect of the safety term. Suppose $\{\lambda_f^*(\rho) , \lambda_d^*(\rho)\}$ is an optimal solution of $ \mathcal P_s(y_0^{(K)}; \cdot)$ with $\rho^{(K)} = \rho$. Then, we can obtain 	
\begin{align}
	J_s(\lambda_f^*(0) , \lambda_d^*(0);0) &\leq J_s(\lambda_f^*(\rho) , \lambda_d^*(\rho);\rho), \nonumber \\
	& \leq J_s(\lambda_f^*(0) , \lambda_d^*(0);0)  \nonumber \\
	& \quad + \rho \cdot \|z^{(K)}_0 - z_f\|  \|\lambda_f^*(0) - \mathbf{1}\|^2, \nonumber \\
	& \leq J_s(\lambda_f^*(0) , \lambda_d^*(0);0) \nonumber \\
	& \quad + \rho (N_{\max}+1) \cdot \|z^{(K)}_0 - z_f\|. \label{error}
\end{align}
To analyze the error derived from the safety term, we need to present an important theorem first, which is demonstrated as Theorem 5.
\begin{theorem}
	\label{thm_sublevel}
	Suppose $\Theta : \mathbb E \to \mathbb R$ is a closed convex function, where $\mathbb E \subseteq \mathbb R^{n_\Theta}$ is a nonempty closed convex set, and suppose the optimal set $X^* := \arg\min_{\mathbf{x} \in \mathbb E}\Theta(\mathbf{x})$ is nonempty and bounded, with minimum value $\Theta_{opt} := \min_{\mathbf{x} \in \mathbb E}\Theta(\mathbf{x})$. For real number $s \ge \Theta_{opt}$, denote by $[\Theta < s]$ and $[\Theta \le s]$ the strict and closed sublevel sets of $\Theta$ at $s$:
	\[\left\{ \begin{gathered}
	[\Theta  < s]: = \left\{ {{\mathbf{x}} \in \mathbb{E}\mid \Theta ({\mathbf{x}}) < s} \right\}, \hfill \\
	[\Theta  \leq s]: = \left\{ {{\mathbf{x}} \in \mathbb{E}\mid \Theta ({\mathbf{x}}) \leq s} \right\}, \hfill \\ 
	\end{gathered}  \right.\]
	and let $\mathrm{Proj}_{X^*}(\mathbf{x}) := \arg\min_{\mathbf{x}' \in X^*} \|\mathbf{x}' - \mathbf{x}\|$ denote the projection of $\mathbf{x}$ onto $X^*$. Then the map $\varphi : [\Theta_{opt},\infty) \to \mathbb R$ defined by
	\[
	\varphi(s):=
	\begin{cases}
	\displaystyle \sup_{\mathbf{x}\in[\Theta<s]} \bigl\| \mathbf{x}-\operatorname{Proj}_{X^{*}}(\mathbf{x}) \bigr\|,
	& s>\Theta_{\mathrm{opt}},\\[4pt]
	0,
	& s=\Theta_{\mathrm{opt}}
	\end{cases}
	\]
	is non-decreasing and upper semi-continuous on $[\Theta_{opt},\infty)$, and
	\begin{equation}
		\lim_{s \to \Theta_{opt}^+} \max_{\mathbf{x} \in [\Theta \le s]}\left\| \mathbf{x} - \mathrm{Proj}_{X^*}(\mathbf{x}) \right\| = 0. \label{sublevel_lim}
	\end{equation}
\end{theorem}
\textbf{\textit{Proof.}} See Appendix~\ref{app:sublevel}. \hfill $\square$

Since $\mathcal P_s(\cdot)$ is convex, combining \eqref{error} with the conclusion of Theorem~\ref{thm_sublevel} shows that the difference between $\{\lambda_f^*(\rho), \lambda_d^*(\rho)\}$ and one of $\{\lambda_f^*(0), \lambda_d^*(0)\}$ converges to zero for sufficiently small $\rho$. Consequently, under these conditions, it is reasonable to regard the solution of $\mathcal P_s(\cdot)$---up to the small effect of the safety term---as a KKT point (i.e., a candidate local optimizer) of $\mathcal P_1'(\cdot)$.

Although the aforementioned strictly convex subproblems are computationally efficient, their local optimality relies on relatively restrictive conditions. For stronger optimality guarantees, SCP serves as a promising alternative to search-based methods by directly solving the original $\mathcal P_1(z_0^{(K)};\cdot)$ instead of its linearized approximation $\mathcal P_1'(y_0^{(K)};\cdot)$. Despite higher computational overhead caused by iterative convexification, converged SCP cycle solutions directly preserve the KKT conditions after trajectory concatenation, yielding valid locally optimal candidates without the assumptions required in Theorem 4, and errors from linearization are typically eliminated. Overall, three strategies are available to address the studied time-optimal control problem: (1) sequential convex solving via the proposed Algorithm 1; (2) direct solution of the original nonconvex problem $\mathcal P_0$ using general nonconvex optimization or SCP; (3) a hybrid scheme that solves $\mathcal P_1(\cdot)$ via standard NLP solvers.
\subsection{Contribution Summary}
This section proposed a receding fixed-horizon framework in which each planning cycle is reformulated into a relaxed convex problem $\mathcal P_r$ and a strict convex problem $\mathcal P_s$ (Subsection 3.1), then solved and concatenated by Algorithm 1 (Subsection 3.2). Two search algorithms realize the framework: CVAPF (Subsection 3.3), an end-to-end trajectory planner robustified by the safety acceleration $a_{safe}$ and the flag $\mathtt{Not\_Safe}$, and CDWA (Subsection 3.4), which adapts the dynamic window approach to nonlinear models. The theoretical guarantees (Subsection 3.5) establish finite termination of Algorithm 1, candidate local-optimality property and convergence of the final time $t_f$ to a locally minimal value as $N_{\max}h \to 0$.
\section{Numerical Experiments}
This section presents numerical experiments that evaluate the proposed algorithm in both static and dynamic environments. All experiments were conducted in MATLAB 2017a on an Intel Core i7 9th-generation processor with 8~GB of RAM. Both algorithms were solved using the CasADi \cite{Casadi} toolbox and the OSQP solver.

\subsection{Static Map Experiments}
\label{sec:static}
In this set of experiments, maps with fixed circular obstacles are generated to evaluate the algorithm performance in static scenarios. The double-integrator dynamics are used, where the constraints \eqref{DQ_1} and \eqref{DQ_2} specialize to
\[
\|z_i[3:4]\| \leq v_{\max}, \qquad \|u_i\| \leq a_{\max}.
\]
The experimental parameters are listed in Table~\ref{tab:static_params}.

\begin{table}[H]
\centering
\caption{Static experiment parameters}
\label{tab:static_params}
\resizebox{\linewidth}{!}{
\begin{tabular}{ccccc}
\toprule\noalign{\smallskip}
Vehicle Parameters & $v_{\max}$(m/s) & $a_{\max}$  & $z_f$ & $z^{(1)}_0$ \\
Range & 12 & 6 & $[160,160,0,0]^\top$ & $0_{4 \times 1}$ \\
\hline\noalign{\smallskip}
Map Parameters & $n_{obs}$ & $r_i$(m) & $\ell$(m) & $\gamma$ \\
Range & 20 & $[3,11]$ & 7 & 3 \\
\noalign{\smallskip}\bottomrule
\end{tabular}}
\end{table}

Obstacle positions and radius are generated using the built-in random number generator, and a validation step ensures target reachability and compliance with the map constraints.

\subsubsection{Single-Map Comparison}
A representative map is selected for detailed comparison. The trajectory planning problem is solved using both Algorithm~1 (with CVAPF) and SCP (parameterized with 35 discretization nodes). Increasing the number of nodes significantly raises the computational cost, while reducing it leads to substantial accuracy loss. The SCP initialization follows a deterministic strategy: $u_i \equiv 0$ and $z_i = (i-1)(z_f - z_0)/N + z_0$; this initial guess is refined by a few SCP iterations with a large trust region.

\begin{figure}[h]
\centering
\includegraphics[width=0.85\linewidth]{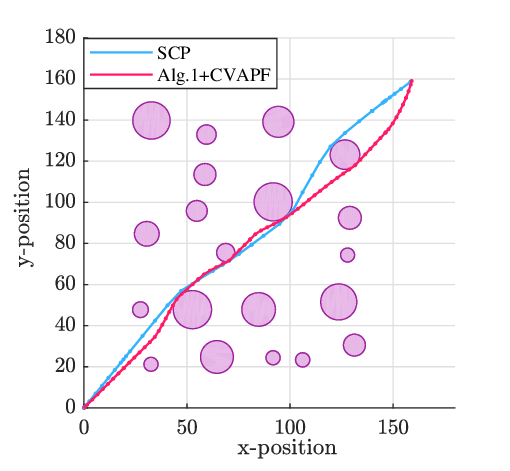}
\caption{Trajectories generated by two distinct algorithms}
\label{fig:fig3}
\end{figure}

The resulting trajectories are shown in Fig.~\ref{fig:fig3}, where the red line corresponds to Algorithm~1 + CVAPF and the blue line to SCP (with $N = 35$). The trajectory generated by Algorithm~1 has 67 nodes (states), which is composed of 14 planning cycles, 12 of which satisfy the preconditions of Theorem~4.

Numerical results are summarized in Table~\ref{tab:quality_single}. Both algorithms achieve comparable final times, while Algorithm~1 reduces the computation time significantly.

\begin{table}[H]
\centering
\caption{Single-map quality comparison}
\label{tab:quality_single}
\begin{tabular}{c c c}
\toprule
& SCP (35 nodes) & Alg.~1 + CVAPF \\
\hline
$t_f$ (s) & 20.92 & 21.12 \\
CT (s) & 1.227 & 0.815 \\
\bottomrule
\end{tabular}
\end{table}

\subsubsection{Batch Statistical Analysis}
\begin{figure}
\centering
\includegraphics[width=1\linewidth]{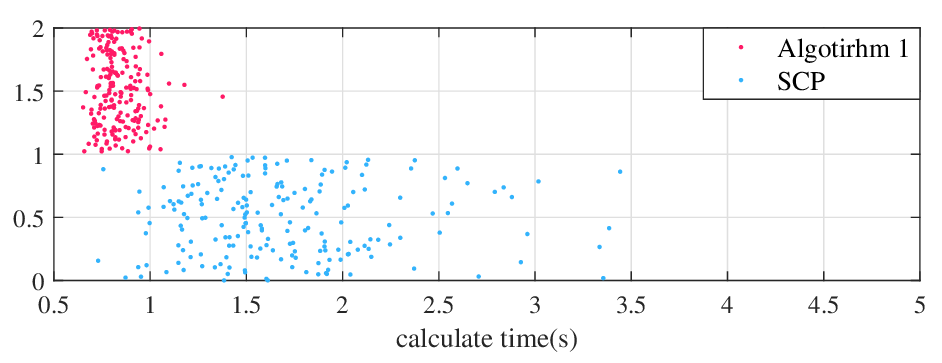}
\caption{Computation time comparison between SCP and Algorithm~1 across 200 static maps}
\label{fig:fig8}
\end{figure}

To further evaluate the algorithm, batch experiments are conducted using the unicycle model with the CDWA search algorithm. Each algorithm is tested on 200 randomly generated maps. The results are shown in Fig.~\ref{fig:fig8} and Table~\ref{tab:batch_static}.

\begin{table}[htbp]
\centering
\caption{Batch static experiment: quality comparison}
\label{tab:batch_static}
\begin{tabular}{c c c}
\toprule
& Alg.~1 + CDWA & SCP \\
\midrule
Success rate (\%) & 90.5 & 87.0 \\
CT average (s) & 0.836 & 1.691 \\
CT worst (s) & 1.377 & 3.442 \\
CT median (s) & 0.814 & 1.597 \\
Average $t_f$ (s) & 21.47 & 21.20 \\
\bottomrule
\end{tabular}
\end{table}

Fig.~\ref{fig:fig8} shows the per-trial computation time of each algorithm across all 200 maps. Algorithm~1 maintained relatively low and concentrated (with standard deviation 0.1044 s) computation times, with most trials completing within 1~s and a median of 0.814~s. In contrast, SCP exhibited higher and more variable computation times (median 1.597~s, standard deviation 0.5097 s), with several trials exceeding 3~s. Moreover, according to these data, Algorithm~1 achieves both a higher success rate and a substantially lower computation time compared with SCP.

The success criteria adopted in these experiments are summarized in Table~\ref{tab:success_criteria}. A trial is counted as successful only if the vehicle reaches the target region without violating any obstacle-avoidance constraint.

\begin{table}[htbp]
\centering
\caption{Success criteria comparison}
\label{tab:success_criteria}
\begin{tabular}{@{}lll@{}}
\toprule
Success criterion & Alg.~1 & SCP \\
\midrule
Solver Feasibility & $\checkmark$ & $\checkmark$ \\
Constraint satisfaction & $\checkmark$ & $\checkmark$ \\
Bounded computation time & $\checkmark$ & $\times$ \\
\bottomrule
\end{tabular}
\end{table}
Solver feasibility for Algorithm~1 requires both connective feasibility (Definition~1) and convex subproblems that are well conditioned and solvable; for SCP, it requires the solvability of the convex subproblems together with the absence of artificial infeasibility arising from the sequential convexification procedure. Constraint satisfaction is assessed through the aggregated-violation criterion
\[\frac{1}{{{N_0} + 1}}\sum\limits_{j = 1}^{{N_0} + 1} {\sum\limits_{k = 1}^{{n_{obs}}} \frac{\max \left( {0,{r_k} - \left\| {{z_j}[1:2] - {p_k}} \right\|} \right)}{\min_k r_k} }  \leq 0.05,\]
where $N_0$ denotes the horizon of the evaluated trajectory, and $r_k$ and $p_k$ denote the radius and the center position of the $k$-th obstacle, respectively. Finally, when computation is online, Algorithm~1 requires each planning cycle to complete within its horizon duration for operational continuity, a bounded per-cycle computation time is also required for a successful trial; this condition does not apply to the offline SCP baseline.

\subsection{Dynamic Map Experiments}
\label{sec:dynamic}
The robustness of Algorithm~1 in time-varying environments is now examined. In this part, we utilize Algorithm~1 in dynamic map environments as a quasi-MPC method\footnote{We use the term \emph{quasi-MPC} to indicate that Algorithm~1 adopts the receding-horizon re-planning paradigm of model predictive control but departs from standard MPC by automatically applying a section of calculated controls rather than constantly applying the first control.}. The experimental parameters are summarized in Table~\ref{tab:dyn_params}.

\begin{table}[!htbp]
\centering
\caption{Parameters for dynamic-map experiments}
\label{tab:dyn_params}
\resizebox{\linewidth}{!}{
\begin{tabular}{ccccc}
\toprule\noalign{\smallskip}
Vehicle Parameters & $v_{\max}$(m/s) & $a_{\max}$(m/s\textsuperscript{2}) & $z_f$ & $z^{(1)}_0$ \\
Range & 12.0 & 6.0 & $[160,160,0,0]^\top$ & $0_{4 \times 1}$ \\
\hline \noalign{\smallskip}
Map Parameters & $n_{\text{obs}}$ & $r_i$(m) & $\ell$(m) & $v_o$(m/s) \\
Range & $\{3,20\}$ & $[3,11]$ & 7 & $0.4\,v_{\max}$ \\
\noalign{\smallskip}\bottomrule
\end{tabular}}
\end{table}
Here $v_o$ denotes the maximum obstacle speed: for each obstacle position $p_i(t)$, $\|p_i(t+\tilde t) - p_i(t)\| \leq v_o \, |\tilde t|$ holds for all $t,\tilde t$. Obstacle motion is simulated using a double-integrator model with saturation. At each step, the current obstacle velocity is measured and a feasible set of accelerations is constructed (analogous to the DWA); an acceleration value is then randomly selected to update the obstacle state.

\subsubsection{Toy Example with Three Obstacles}
\begin{figure}[h]
\centering
\subfloat[Perspective 1]{
\includegraphics[scale=0.51]{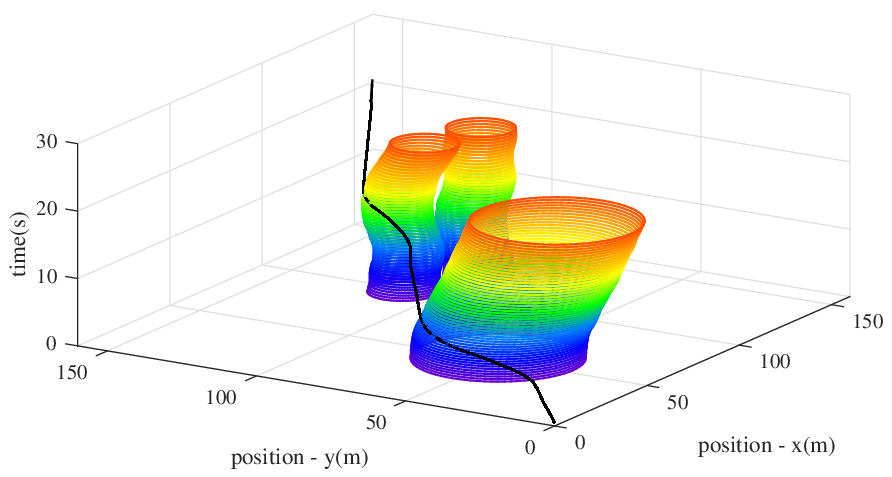}}\\
\subfloat[Perspective 2]{
\includegraphics[scale=0.51]{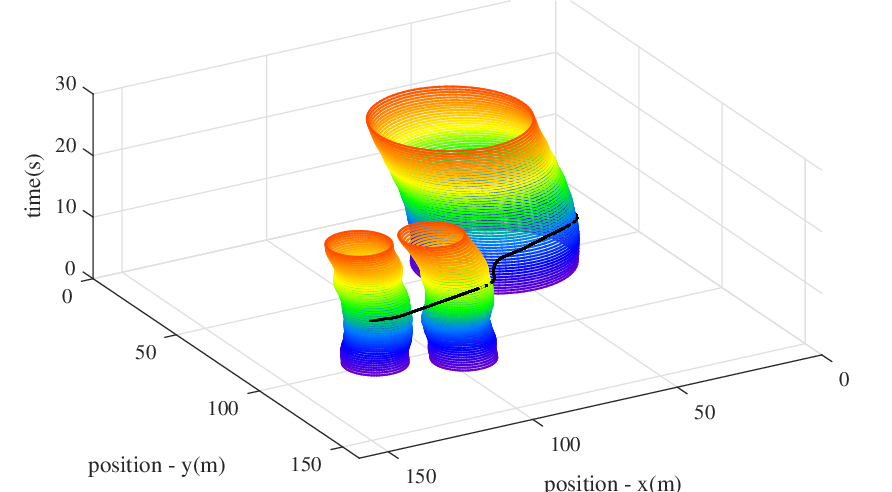}}
\caption{Obstacle avoidance in a dynamic environment with three moving obstacles}
\label{fig:3d_traj}
\end{figure}

A sparse configuration with three moving obstacles is first considered. The resulting trajectory is visualized from two perspectives in Fig.~\ref{fig:3d_traj}, where the vertical axis represents time and each horizontal slice indicates the obstacle danger zones at the corresponding moment.

\begin{figure}[!h]
\centering
\includegraphics[width=1.02\linewidth]{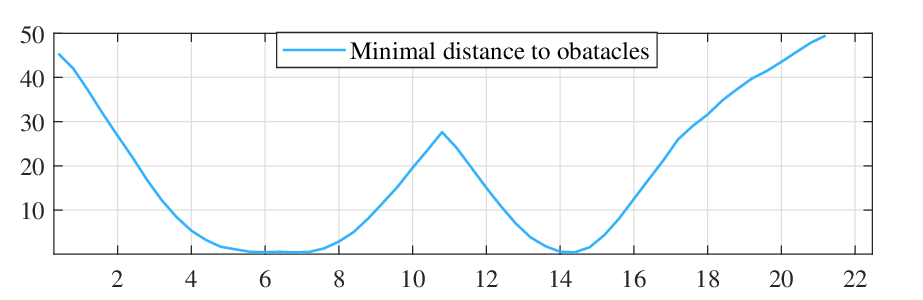}
\caption{Minimum distance to danger zones during the dynamic task}
\label{fig:min_dist}
\end{figure}

The black line in each subplot traces the vehicle path. Fig.~\ref{fig:min_dist} plots the minimum distance between the vehicle and any danger zone throughout the task. A clearance of at least 0.4212~m is maintained at all times.

\subsubsection{Batch Statistical Analysis}
To assess scalability, experiments with $n_{\text{obs}} = 20$ moving obstacles across 200 randomly generated maps are conducted. Both CVAPF and CDWA search algorithms are evaluated. The results are reported in Table~\ref{tab:batch_dyn}.

\begin{table}[!htbp]
	\centering
	\caption{Batch dynamic experiment: quality comparison}
	\label{tab:batch_dyn}
	\small
	\begin{tabular}{lcc}
		\toprule
		Metric & Alg.~1 + CVAPF & Alg.~1 + CDWA \\
		\midrule
		Succ. (\%) & 88.5 & 89.0 \\
		CT1 avg. (ms) & 58.7 & 69.3 \\
		CT1 med. (ms) & 59.2 & 71.1 \\
		\bottomrule
	\end{tabular}
	\small{CT1: the average computation time required to generate one second of the state-input sequence within a single task.
	}
\end{table}

These results demonstrate that the proposed planning-cycle algorithm reliably produces collision-free trajectories in both static and dynamic environments, with computation times significantly lower than those of the SCP baseline.
\section{Conclusion}
The proposed planning-cycle framework achieves a higher success rate and substantially lower computation time than sequential convex programming in both static and dynamic environments, at comparable control time. The optimality claim is deliberately qualified: as defined in the Introduction, ``near time-optimal'' means that the generated trajectories are candidate optimal, in the discretized time dimension, with a small deviation that stems from factors inherent to any discrete-time, model-based planner.

Several limitations remain. Feasibility is guaranteed only at discrete time points, so interpolated and simulated states may violate constraints. Excessively long planning horizons compromise solution timeliness in dynamic maps, and a planning cycle whose computation time exceeds the horizon of the latest control sequence may cause temporary loss of vehicle control in real-time dynamic-map applications. Future work will focus on finding more reliable trajectory search methods, establishing a more comprehensive theoretical analysis of convexification equivalence, and enhancing the robustness of Algorithm 1.

\bibliographystyle{unsrt}        
\bibliography{autosam}           

\appendix
\section{Proof of Theorem 2}

We first record the KKT conditions of $\mathcal P_1(z'_0; M)$ for a general horizon $M$, with $z'_0$ the reference state. Suppose $\{z_1 ^*, z_2 ^*, \cdots, z_{M+1}^*; u_1^*, \cdots, u_M^*\}$ is a local optimizer of $\mathcal P_1(z'_0; M)$ with Lagrange multipliers $\upsilon_i^*, \zeta_{ij}^*, s_i^*, q_i^*$. Recall from the remark below \eqref{new_init} that $z_1^*$ is pinned to $z'_0$ and is not a decision variable, so that no stationarity condition is imposed at $i = 1$. By Assumption~\ref{KKT}, the following conditions hold:
\begin{subequations}
	\begin{align}
		&\begin{aligned}
			\frac{{\partial \mathscr L_M(z'_0;\cdot)}}{{\partial z_i^*}} &= \frac{\partial}{\partial z_i^*} {\sum\limits_{j = 1}^{{n_{obs}}} {{\zeta_{ij}^*}\left( {r_j^2 - {{\left\| {z_i^*[1:2] - {p_j}} \right\|}^2}} \right)} }   \\
			& \quad  - {\frac{\partial f}{\partial z_i^*}}^\top \upsilon_i^* + \upsilon_{i - 1}^* + 2s_i^* {Q_1}z_i^* \\
			& = \mathbf 0,
		\end{aligned} \label{th2mna}\\
		&\frac{{\partial \mathscr L_M(z'_0;\cdot)}}{{\partial u_i^*}} = - {\frac{\partial f}{\partial u_i^*}}^\top \upsilon_i^* + 2 q_i^* {Q_2}u_i^* = \mathbf 0, \label{th2mnb}\\
		&\upsilon_i^*\left( {z_{i + 1}^*} - f(z_i^*, u_i^*) \right) = \mathbf 0, \label{th2mnc}\\
		&{\zeta _{ij}^*}\left( {r_j^2 - {{\left\| {z_i^*[1:2] - {p_j}}\right\|}^2}} \right) = \mathbf 0, \label{th2mnd}\\
		&s_i^*{G_1}(z_i^*) = 0, \label{th2mne}\\
		&q_i^*{G_2}(u_i^*) = 0, \label{th2mnf}\\
		& \zeta_{ij}^*, q_i^*, s_i^* \geq 0, \label{th2mng}
	\end{align}
\end{subequations}
where $\upsilon_{M+1}^* = \mathbf 0$, the range of $i$ is $1 \leq i \leq M$ in \eqref{th2mnb}, \eqref{th2mnc}, \eqref{th2mnf}, $2 \leq i \leq M$ in \eqref{th2mna}, $1 \leq i \leq M+1$ in \eqref{th2mnd}, \eqref{th2mne}, and the terminal stationarity condition, which replaces \eqref{th2mna} at $i = M+1$, is
\begin{align}
	\frac{{\partial \mathscr L_M(z'_0;\cdot)}}{{\partial z_{M+1}^*}} &= \frac{\partial}{\partial z_{M+1}^*} \sum\limits_{j = 1}^{{n_{obs}}} {\zeta_{M+1,j}^*}\Big( r_j^2 \nonumber \\
	& \quad - {\left\| {z_{M+1}^*[1:2] - {p_j}} \right\|}^2 \Big) \nonumber \\
	& \quad - {\frac{\partial f}{\partial z_{M+1}^*}}^\top\upsilon_{M+1}^* + {\upsilon_{M}^*} \nonumber \\
	& \quad + 2 s_{M+1}^* {Q_1}z_{M+1}^* + 2Q_0(z_{M+1}^* - z_f) \nonumber \\
	& = \mathbf 0. \label{th2mnh}
\end{align}
Applying \eqref{th2mna}-\eqref{th2mnh} with $M = m$ to the optimizer $\{z_1^1, z_2^1, \cdots, z_{m+1}^1; u_1^1, \cdots, u_m^1\}$ of $\mathcal P_1(z'_0; m)$, and with $M = n$ to the optimizer $\{z_1^2, z_2^2, \cdots, z_{n+1}^2; u_1^2, \cdots, u_n^2\}$ of $\mathcal P_1(z_{m'+1}^1; n)$, yields multipliers $\upsilon_i^\ell, \zeta_{ij}^\ell, q_i^\ell, s_i^\ell$ ($\ell = 1,2$) satisfying \eqref{th2mna}-\eqref{th2mng} under the natural index relabeling, with $\upsilon_{m+1}^1 = \upsilon_{n+1}^2 = \mathbf 0$ and the terminal condition \eqref{th2mnh} holding at $z_{m+1}^1$ and $z_{n+1}^2$.
Now concatenate the two trajectories by setting $z_i := z_i^1$ ($1 \leq i \leq m'+1$), $z_{m'+1+j} := z_{j+1}^2$ ($1 \leq j \leq n$), $u_i := u_i^1$ ($1 \leq i \leq m'$), $u_{m'+j} := u_j^2$ ($1 \leq j \leq n$), and define the concatenated multipliers
\begin{align*}
	\bm{\zeta} &:= \big\{\zeta_{1j}^1, \zeta_{2j}^1, \cdots, \zeta_{m',j}^1, \zeta_{1j}^2, \zeta_{2j}^2, \cdots, \zeta_{n+1,j}^2\big\}_{j = 1}^{n_{obs}},\\
	\bm{\upsilon} &:= \big\{ \upsilon_1^1, \cdots, \upsilon_{m'}^1, \upsilon_{1}^2, \cdots, \upsilon_{n}^2 \big\},\\
	\mathbf q &:= \{q_1^1, \cdots, q_{m'}^1, q_1^2, \cdots, q_{n}^2 \},\\
	\mathbf s &:= \{s_1^1, \cdots, s_{m'}^1, s_1^2, s_2^2, \cdots, s_{n+1}^2 \}.
\end{align*}
where $\upsilon_{m'+n+1} = \upsilon_{n+1}^2 = \mathbf 0$. By \eqref{th2mna}-\eqref{th2mnh} applied to each subproblem, the concatenated sequence satisfies every KKT condition of $\mathcal P_1(z'_0; m'+n)$ except possibly the stationarity of the Lagrangian at the stitching state $z_{m'+1} = z_1^2$: primal feasibility, the complementarity and sign conditions are inherited index by index from the two subproblems, and the stationarity with respect to $u_i$ and to $z_i$ ($i \neq m'+1$) and the terminal stationarity at $z_{m'+n+1} = z_{n+1}^2$ follow from the corresponding conditions of the two subproblems. The one condition that is not inherited is the stationarity at the stitching state, a genuine requirement because $m'+1 \geq 2$ makes $z_{m'+1}$ a decision variable of $\mathcal P_1(z'_0; m'+n)$, i.e.,
\begin{align}
	&\frac{\partial }{{\partial {z_{m'+1}}}}\sum\limits_{j = 1}^{{n_{obs}}} {{\zeta_{m'+1,j}}\left( {r_j^2 - {{\left\| {z_{m'+1}[1:2] - {p_j}} \right\|}^2}} \right)} \nonumber \\
	& \quad - {\frac{\partial f}{\partial z_{m'+1}}}^\top \upsilon_{m'+1} + \upsilon_{m'} \nonumber \\
	& \quad + 2s_{m'+1}Q_1 z_{m'+1} = \mathbf 0. \label{th2_3}
\end{align}
It remains to verify \eqref{th2_3}. Since the stitching state $z_{m'+1}^1 = z_1^2$ is strictly feasible for the collision and state-bound constraints, the complementarity conditions \eqref{th2mnd} and \eqref{th2mne} at $i = 1$ of the second subproblem give $\zeta_{1j}^2 = 0$ for all $j$ and $s_1^2 = 0$. Under the multiplier mapping $z_{1}^2 = z_{m'+1}$, $\zeta_{1,j}^2 = \zeta_{m'+1,j}$, $s_{1}^2 = s_{m'+1}$, $\upsilon_{1}^2 = \upsilon_{m'+1}$ and $\upsilon_{m'}^1 = \upsilon_{m'}$, the first and the last terms of \eqref{th2_3} vanish, so that \eqref{th2_3} reduces to the single condition
\begin{equation}
	\upsilon_{m'}^1 = {\frac{\partial f}{\partial z_{m'+1}}}^\top \upsilon_1^2, \label{th2_junction}
\end{equation}
in which the Jacobian is evaluated at $(z_1^2, u_1^2)$. It remains to establish \eqref{th2_junction}, and we do so by showing that both transmitted costates vanish. Write $A_i^\ell := {\partial f}/{\partial z_i^\ell}$ and $B_i^\ell := {\partial f}/{\partial u_i^\ell}$ for $\ell = 1,2$, and set $\bar m := m'+\tau-1$.

Since ${u_i^1}^\top Q_2 u_i^1 < 1$ for $m' \leq i \leq \bar m$ and ${u_i^2}^\top Q_2 u_i^2 < 1$ for $1 \leq i \leq \tau$, complementarity \eqref{th2mnf} gives $q_i^1 = 0$ and $q_i^2 = 0$ on these ranges, so that \eqref{th2mnb} reduces to
\begin{align*}
	\begin{cases}
		{B_i^1}^\top \upsilon_i^1 = \mathbf 0, & m' \leq i \leq \bar m, \\[4pt]
		{B_i^2}^\top \upsilon_i^2 = \mathbf 0, & 1 \leq i \leq \tau.
	\end{cases}
\end{align*}
Likewise, the states $z_i^1$ ($m'+1 \leq i \leq \bar m$) and $z_i^2$ ($1 \leq i \leq \tau$) are strictly feasible for the collision and state-bound constraints, so \eqref{th2mnd} and \eqref{th2mne} give $\zeta_{ij}^\ell = 0$ and $s_i^\ell = 0$ on these ranges, and \eqref{th2mna} reduces to the costate recursions
\begin{equation*}
	\left\{
	\begin{aligned}
		\upsilon_{i-1}^1 &= {A_i^1}^\top \upsilon_i^1, && m'+1 \leq i \leq \bar m, \\[4pt]
		\upsilon_{i-1}^2 &= {A_i^2}^\top \upsilon_i^2, && 2 \leq i \leq \tau.
	\end{aligned}
	\right.
\end{equation*}
which are well defined because $m'+1 \geq 2$, $\bar m \leq m$ and $\tau \leq n$. Propagating each recursion to the end of its window yields $\upsilon_j^1 = {\left( A_{j+1}^1 \cdots A_{\bar m}^1 \right)}^\top \upsilon_{\bar m}^1$ for $m' \leq j \leq \bar m$ and $\upsilon_j^2 = {\left( A_{j+1}^2 \cdots A_\tau^2 \right)}^\top \upsilon_\tau^2$ for $1 \leq j \leq \tau$ (empty products being the identity). Substituting these into ${B_j^\ell}^\top \upsilon_j^\ell = \mathbf 0$ and stacking over $j$ gives
\begin{align}
	{\left[ B_{\bar m}^1, \cdots, A_{\bar m}^1 \cdots A_{m'+1}^1 B_{m'}^1 \right]}^\top \upsilon_{\bar m}^1 &= \mathbf 0, \nonumber \\
	{\left[ B_\tau^2, \cdots, A_\tau^2 \cdots A_{2}^2 B_{1}^2 \right]}^\top \upsilon_\tau^2 &= \mathbf 0, \label{th2_reach}
\end{align}
whose coefficient matrices are precisely the $\tau$-step reachability matrices \eqref{reach_matrix} formed along the two optimizers. By Assumption~\ref{Full_Rank} both have full row rank $n_z$, hence $\upsilon_{\bar m}^1 = \upsilon_\tau^2 = \mathbf 0$, and the recursions above then give $\upsilon_{m'}^1 = \mathbf 0$ and $\upsilon_1^2 = \mathbf 0$. Condition \eqref{th2_junction} is therefore satisfied. Consequently every KKT condition of $\mathcal P_1(z'_0; m'+n)$ is satisfied by the concatenated sequence, which completes the proof. \hfill $\square$

\section{Proof of Theorem 4}
For simplicity, we omit the iteration superscripts ($z^{(K)}$ is abbreviated as $z$) and take $\rho = 0$. The Lagrange function of $\mathcal P_1'(y_0; N; \cdot)$ is $\mathscr L_N(y_0; z,u,\upsilon,\zeta,s,q)$, of the same form as \eqref{lagrange}. Under Assumption~\ref{KKT}, a local optimizer with the corresponding multipliers satisfies the KKT conditions of $\mathcal P_1'(\cdot)$: the primal feasibility and complementarity conditions of the template \eqref{lagrange} together with the stationarity condition
\begin{equation}
	\{ {z^*_i},{u^*_i}\}  = \mathop {\arg \min }\limits_{{z _{i}},{u _{i}}} {\mathscr L_N} \label{argmin}
\end{equation}
for proper indices $i$. We show that the state-input sequence generated from the optimizer of $\mathcal P_s(y_0; \tilde{\mathbf y}_N; \bar{\mathbf y}_N; \cdot)$ satisfies these conditions. The Lagrange function of $\mathcal P_s(\cdot)$, $\mathscr L_s(z_0; z,u,\upsilon,\eta,\sigma,\xi,\mu,s,q)$, is obtained from $\mathscr L_N$ by deleting the obstacle terms and adjoining the multipliers of the $\lambda$-bounds:
\begin{align}
	\mathscr L_s & = \mathscr L_N \nonumber \\
	& \quad - \sum\limits_{i = 1}^{N + 1} {\sum\limits_{j = 1}^{{n_{obs}}} {{\zeta _{ij}}\left( {r_j^2 - {{\left\| {{z_i}[1:2] - {p_j}} \right\|}^2}} \right)} } \nonumber \\
	& \quad + \sum\limits_{i = 1}^{N + 1} \Big[({\eta_i} - {\sigma_i}){\lambda _{f,i}} - {\eta_i} \nonumber \\
	& \quad + {\xi_i}({\lambda _{d,i}} - {k_i}{\lambda _{f,i}}) - {\mu_i}{\lambda _{d,i}}\Big],
\end{align}
where $\eta_i, \sigma_i, \mu_i, \xi_i$ are the Lagrange multipliers of the constraints $\lambda_{f,i} \le 1$, $\lambda_{f,i} \ge 0$, $\lambda_{d,i} \ge 0$, and $\lambda_{d,i} \le k_i \lambda_{f,i}$, respectively.

Any feasible point of $\mathcal P_s(\cdot)$, transferred via \eqref{z_transfer}, automatically satisfies \eqref{c2d_var}, \eqref{new_init}, \eqref{DQ_1}, \eqref{DQ_2}; hence it is feasible for $\mathcal P_1'(\cdot)$ precisely when it is collision-free, which the convex region $\mathcal Z_i$ guarantees by construction. The two conditions of Theorem~4 certify this collision-free property together with the stationarity, complementarity, and dual feasibility of the KKT system of $\mathcal P_1'(\cdot)$.

\emph{Case 1} ($\lambda_{\min,i} = 0$ for all $i$). Then $\bar z_i$ is collision-free for every $i$, so $\bar{\mathbf y}_N$ is feasible for $\mathcal P_1'(\cdot)$. Since $\mathcal P_r(\cdot)$ is $\mathcal P_1'(\cdot)$ with only the non-convex collision constraints \eqref{nonconvex_position_con_discrete} removed, its optimizer $\bar{\mathbf y}_N$ attains $J(\bar{\mathbf y}_N) \le J( \mathbf y)$ for every $ \mathbf y$ feasible for $\mathcal P_1'(\cdot)$; as $\bar{\mathbf y}_N$ itself is feasible, it is a \emph{global} optimizer of $\mathcal P_1'(\cdot)$. The alternative $\tilde z_i = \bar z_i$ for all $i$ is analogous: $\tilde{\mathbf y}_N$ is feasible for $\mathcal P_1'(\cdot)$ and has the same state sequence, so $\bar{\mathbf y}_N$ is feasible as well. In either alternative, $\lambda_f = \lambda_d = {\bf 0}$ is feasible for $\mathcal P_s(\cdot)$ and generates $\bar{\mathbf y}_N$ via \eqref{z_transfer}, so the optimal value of $\mathcal P_s(\cdot)$ is no larger than $J(\bar{\mathbf y}_N)$; conversely, every feasible point of $\mathcal P_s(\cdot)$ satisfies the constraints of $\mathcal P_r(\cdot)$, so the optimal value of $\mathcal P_s(\cdot)$ is no smaller than $J(\bar{\mathbf y}_N)$. The two values coincide, and every optimizer of $\mathcal P_s(\cdot)$, transferred to $(z,u)$, is collision-free by construction of $\mathcal Z_i$, attains $J(\bar{\mathbf y}_N)$, and is thus a global optimizer of $\mathcal P_1'(\cdot)$; under Assumption~\ref{KKT} it satisfies the KKT conditions of $\mathcal P_1'(\cdot)$.

\emph{Case 2} (relative interior solution). By Definition~\ref{def_rel_int}, every inequality constraint of $\mathcal P_s(\cdot)$ is strict at the optimizer: $0 < \lambda_{d,i} < \lambda_{f,i} < 1$, $0 \leq \lambda_{d,i} < k_i \lambda_{f,i}$ for all $i, j$, $G_1(z_i) < 0$ for all $i$, and $G_2(u_i) < 0$ for all $i \le N$; hence the pair is feasible for $\mathcal P_1'(\cdot)$. Since $\mathcal P_s(\cdot)$ is convex and its optimizer lies in the relative interior, no inequality is active, so complementary slackness gives $\zeta_{ij}^* = \eta_i^* = \sigma_i^* = \mu_i^* = \xi_i^* = s_i^* = q_i^* = 0$ for all $i,j$; the stationarity condition below then holds without any constraint qualification, and $\mathscr L_s \equiv \mathscr L_N$. The first-order optimality conditions of $\mathcal P_s(\cdot)$ reduce to
\begin{equation}
	\{ {\lambda _{f,i}^*},{\lambda _{d,i}^*}\}  = \mathop {\arg \min }\limits_{{\lambda _{f,i}},{\lambda _{d,i}}} {\mathscr L_s}. \label{argmin_lambda}
\end{equation}

In $\mathcal P_s(\cdot)$, \eqref{z_transfer} and the nonparallel anchor directions make the map $\{\lambda_{f,i},\lambda_{d,i}\} \mapsto \{z_1, \cdots, z_{N+1}\}$ a bijection onto the region $\mathcal R$, and \eqref{c2d_var} together with Assumption~\ref{Full_Rank} determines the controls from the states, $u_i = (B^\top B)^{-1}B^\top(z_{i+1} - A z_i - w)$. With $\mathscr L_s \equiv \mathscr L_N$, \eqref{argmin_lambda} is the first-order condition of $\mathscr L_N$ w.r.t.\ the interpolation variables, $\partial \mathscr L_N/\partial \lambda_{f,i}^* = \partial \mathscr L_N/\partial \lambda_{d,i}^* = 0$ for all $i$; by the bijection, the state-input pair is a minimizer of $\mathscr L_N$ over the region $\mathcal R$. Moreover $\mathscr L_N$ is convex in $(z,u)$: the obstacle terms in \eqref{lagrange} vanish ($\zeta_{ij}^* = 0$), the terminal cost \eqref{J_P_1} and the bounds $G_1, G_2$ ($s_i, q_i \geq 0$) are convex, and the dynamics terms are affine. Since the relative interior solution lies strictly inside $\mathcal R$, the pair is an interior minimizer of the convex function $\mathscr L_N$ over $\mathcal R$; its first-order conditions along every direction of $\mathcal R$ hold, which is the KKT stationarity condition \eqref{argmin} of $\mathcal P_1'(\cdot)$ within $\mathcal R$. Together with the primal feasibility, complementarity, and dual feasibility above, the computed pair satisfies the KKT conditions of $\mathcal P_1'(\cdot)$ within the region $\mathcal R$; in Case~1 the conclusion holds for the full problem by global optimality under Assumption~\ref{KKT}. This completes the proof. \hfill $\square$

\section{Proof of Theorem \ref{thm_sublevel}}\label{app:sublevel}

We first record two preliminary facts used throughout the proof.

\textbf{1.Compactness of sublevel sets} For every $s \ge \Theta_{opt}$ the set $[\Theta \le s]$ is nonempty (it contains $X^*$), closed, and convex, and it is bounded because its recession cone is trivial. Indeed, if $d$ were a nonzero recession direction, then for any $\mathbf{x}^* \in X^*$ the ray $\mathbf{x}^* + t\,d$, $t \ge 0$, lies in $[\Theta \le s]$, so the convex function $g(t) := \Theta(\mathbf{x}^* + t\,d)$ satisfies $g(t) \le s$ and $g(0) = \Theta_{opt}$. For $0 \le a \le t$, convexity gives
\begin{align}
	g(a) &\le \left(1 - \frac{a}{t}\right)\Theta_{opt} + \frac{a}{t}\,g(t) \nonumber \\ 
	&\le \left(1 - \frac{a}{t}\right)\Theta_{opt} + \frac{a}{t}\,s. \nonumber
\end{align}
Letting $t \to \infty$ yields $g(a) \le \Theta_{opt}$; since $\Theta_{opt}$ is the global minimum of $\Theta$, we get $g(a) = \Theta_{opt}$ for every $a \ge 0$, so the whole ray is contained in $X^*$, contradicting the boundedness of $X^*$. Hence $[\Theta \le s]$ is compact for every $s \ge \Theta_{opt}$.

\textbf{2. Closure identity} For $s > \Theta_{opt}$ we have $\mathrm{cl}\,[\Theta < s] = [\Theta \le s]$. The inclusion $\subseteq$ holds because $[\Theta \le s]$ is closed and contains $[\Theta < s]$. Conversely, let $\mathbf{x} \in [\Theta \le s]$. If $\Theta(\mathbf{x}) < s$ there is nothing to prove; if $\Theta(\mathbf{x}) = s$, fix $\mathbf{x}^* \in X^*$. The segment $\mathbf{z}_\lambda := (1-\lambda)\mathbf{x}^* + \lambda\,\mathbf{x}$, $\lambda \in [0,1]$, lies in $\mathbb E$ (convexity of $\mathbb E$) and satisfies, by the convexity of $\Theta$,
\begin{equation}
	\Theta(\mathbf{z}_\lambda) \le (1-\lambda)\,\Theta_{opt} + \lambda\,s < s, \quad \lambda < 1, \nonumber
\end{equation}
so $\mathbf{z}_\lambda \in [\Theta < s]$ and $\mathbf{z}_\lambda \to \mathbf{x}$ as $\lambda \to 1$; hence $\mathbf{x} \in \mathrm{cl}\,[\Theta < s]$.

Since $X^*$ is a nonempty closed convex set, the projection $\mathrm{Proj}_{X^*}(\mathbf{x})$ is the unique minimizer of $\|\mathbf{y} - \mathbf{x}\|$ over $X^*$, and the map $\mathbf{x} \mapsto \mathrm{Proj}_{X^*}(\mathbf{x})$ is non-expansive, hence continuous. By Steps 1 and 2, $[\Theta \le s]$ is compact and equals $\mathrm{cl}\,[\Theta < s]$ for $s > \Theta_{opt}$, so the continuous map $\mathbf{x} \mapsto \|\mathbf{x} - \mathrm{Proj}_{X^*}(\mathbf{x})\|$ attains its maximum over $[\Theta \le s]$, and this maximum equals the supremum over the open strict sublevel set:
\begin{align}
	\varphi(s) &= \sup_{\mathbf{x} \in [\Theta < s]}\left\| \mathbf{x} - \mathrm{Proj}_{X^*}(\mathbf{x}) \right\| \nonumber \\
	 &= \max_{\mathbf{x} \in [\Theta \le s]}\left\| \mathbf{x} - \mathrm{Proj}_{X^*}(\mathbf{x}) \right\|, \label{phi_max}
\end{align}
where $s > \Theta_{opt}$. Indeed, the maximum over the compact set $[\Theta \le s] = \mathrm{cl}\,[\Theta < s]$ is attained at a point that is a limit of points of $[\Theta < s]$, so it does not exceed the supremum over $[\Theta < s]$, and the reverse inequality is immediate. At the endpoint, $[\Theta \le \Theta_{opt}] = X^*$ and \eqref{phi_max} reduces to $\varphi(\Theta_{opt}) = 0$. Monotonicity of $\varphi$ follows from the nesting of the sublevel sets.

It remains to prove upper semi-continuity and the limit. Let $s_n \to s_0 \ge \Theta_{opt}$; for each $n$ with $s_n > \Theta_{opt}$ choose $\mathbf{x}_n \in [\Theta \le s_n]$ attaining the maximum in \eqref{phi_max}, and for $s_n = \Theta_{opt}$ take any $\mathbf{x}_n \in X^*$ (so that $\varphi(s_n) = \|\mathbf{x}_n - \mathrm{Proj}_{X^*}(\mathbf{x}_n)\|$ holds in either case). All $[\Theta \le s_n]$ are contained in the compact set $[\Theta \le \bar s]$ with $\bar s := \sup_n s_n < \infty$, so a subsequence $\mathbf{x}_{n_k}$ converges to some $\mathbf{x}_*$. Since $\Theta$ is closed, hence lower semi-continuous, and $\Theta(\mathbf{x}_{n_k}) \le s_{n_k} \to s_0$, we have $\Theta(\mathbf{x}_*) \le s_0$, i.e., $\mathbf{x}_* \in [\Theta \le s_0]$. Consequently,
\begin{align}
	\lim_{k \to \infty}\varphi(s_{n_k}) &= \lim_{k \to \infty}\left\| \mathbf{x}_{n_k} - \mathrm{Proj}_{X^*}(\mathbf{x}_{n_k}) \right\|, \nonumber \\
			&= \left\| \mathbf{x}_* - \mathrm{Proj}_{X^*}(\mathbf{x}_*) \right\| \le \varphi(s_0), \nonumber
\end{align}
which proves that $\varphi$ is upper semi-continuous at every $s_0 \ge \Theta_{opt}$. Finally, applying the same argument to a sequence $s_n \downarrow \Theta_{opt}$ gives $\mathbf{x}_* \in [\Theta \le \Theta_{opt}] = X^*$, hence $\left\| \mathbf{x}_* - \mathrm{Proj}_{X^*}(\mathbf{x}_*) \right\| = 0$ and therefore $\lim_{s \to \Theta_{opt}^+}\varphi(s) = 0$. Since $\mathrm{cl}\,[\Theta < s] = [\Theta \le s]$ for $s > \Theta_{opt}$ by Step 2 and the maximum in \eqref{phi_max} is attained, this is exactly \eqref{sublevel_lim}. This completes the proof. \hfill $\square$
\end{document}